\documentclass[letterpaper]{article} 
\usepackage{aaai23}  
\nocopyright  
\usepackage{times}  
\usepackage{helvet}  
\usepackage{courier}  
\usepackage[hyphens]{url}  
\usepackage{graphicx} 
\usepackage{natbib}  
\usepackage{caption} 
\usepackage{algorithm}
\usepackage{algorithmic}

\usepackage{newfloat}
\usepackage{listings}
\DeclareCaptionStyle{ruled}{labelfont=normalfont,labelsep=colon,strut=off} 
\floatstyle{ruled}
\newfloat{listing}{tb}{lst}{}
\floatname{listing}{Listing}
\usepackage[utf8]{inputenc}
\usepackage{CJKutf8}
\usepackage{subcaption}
\usepackage{xcolor}
\usepackage{amsmath}
\usepackage{amssymb}
\usepackage{diagbox}
\usepackage{cleveref}
\usepackage{multirow}
\usepackage{booktabs}
\usepackage[normalem]{ulem} 

\title{GenéLive! Generating Rhythm Actions in Love Live!}
\author{
    Atsushi Takada\textsuperscript{\rm 1},
    Daichi Yamazaki\textsuperscript{\rm 1},
    Yudai Yoshida\textsuperscript{\rm 1},
    Nyamkhuu Ganbat\textsuperscript{\rm 1},
    Takayuki Shimotomai\textsuperscript{\rm 1},
    Naoki Hamada\textsuperscript{\rm 1},
    Likun Liu\textsuperscript{\rm 2},
    Taiga Yamamoto\textsuperscript{\rm 2},
    Daisuke Sakurai\textsuperscript{\rm 2}
}
\affiliations{
    \textsuperscript{\rm 1}KLab Inc.\\
    \textsuperscript{\rm 2}Kyushu University\\

    \{takada-at,yamazaki-d, yoshida-yud, ganbat-n, shimotomai-t, hamada-n\}@klab.com,\\
    \{liu.likun.654, yamamoto.taiga.160\}@s.kyushu-u.ac.jp,
    d.sakurai@ieee.org
}

\begin{document}
\maketitle

\begin{abstract}
This article presents our generative model for \emph{rhythm action games} together with applications in business operation.
Rhythm action games are video games in which the player is challenged to issue commands at the right timings during a music session.
The timings are rendered in the \emph{chart}, which consists of visual symbols, called \emph{notes}, flying through the screen.
We introduce our deep generative model, \emph{GenéLive!}, which outperforms the state-of-the-art model by taking into account musical structures through beats and temporal scales.
Thanks to its favorable performance, GenéLive!~was put into operation at KLab Inc., a Japan-based video game developer, and reduced the business cost of chart generation by as much as half.
The application target included the phenomenal ``Love Live!,'' which has more than 10 million users across Asia and beyond, and is one of the few rhythm action franchises that has led the online-era of the genre.
In this article, we evaluate the generative performance of GenéLive!~using production datasets at KLab as well as open datasets for reproducibility, while the model continues to operate in their business.
Our code and the model, tuned and trained using a supercomputer, are publicly available.
\end{abstract}

\section{Introduction}\label{sec:introduction}
The success of deep generative models is rapidly spreading over the entire fields of industry and academia.
In today's game developments, deep generative models are starting to help us create various assets including graphics, sounds, character motions, conversations, landscapes, and level designs.
For instance, the Game Developers Conference 2021\footnote{\url{https://gdconf.com/}.} held a special session named ``Machine Learning Summit'' to present various deep generative models used in game products, such as for generating character motions that match the content of conversations \citep{YuDing2021} and for generating 3D facial expression models of characters from human face pictures \citep{PeiLi2021}.

\begin{figure}[tb]
    \centering
    \includegraphics[width=\linewidth,clip,trim=0 60 0 60]{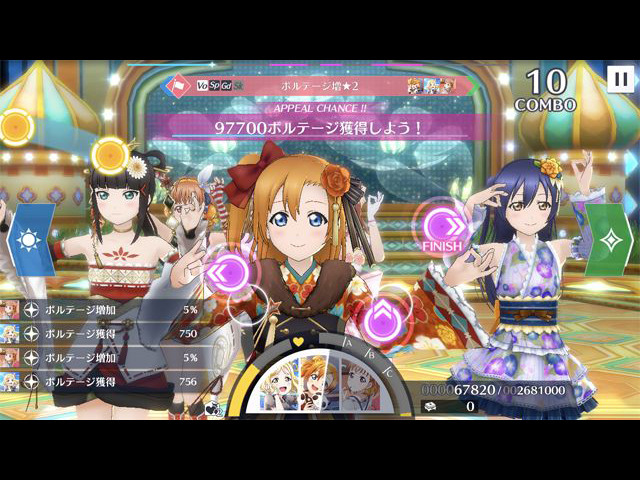}
    \caption{Love Live!~School Idol Festival All Stars.\protect\footnotemark
    }
    \label{fig:game_screen}
\end{figure}
\footnotetext{©2013 PROJECT Lovelive! ©2017 PROJECT Lovelive!~Sunshine!! ©2020 PROJECT Lovelive!~Nijigasaki High School Idol Club ©Bushiroad International ©SUNRISE ©bushiroad}

Although deep generative models for rhythm actions have been studied for a while -- notably by \citet{Donahue:2017}, they have been focusing on proof of concept or personal hobby use, not yet being used in cutting-edge commercial products.
There thus remain questions: what is the blocker to leverage chart generation models in the game business, and how should we overcome it?

The present article reveals a key remaining problem, which is musical structure recognition.
Indeed, we considered features such as beats and temporal scales with our model (see \cref{sec:musicalStructure} for more on musical structures).
While existing models have not had components dedicated for these concepts,
our preset model has them, which process beats and multiple temporal scales.
Those correspond to our \emph{beat guide} and \emph{multi-scale conv-stack}, respectively.

The beat guide is an extraordinary technique in the sense that it can be computed for any input audio and straightforwardly.
Somewhat surprisingly, it had been overlooked by existing works.
The multi-scale conv-stack is incorporated in order to capture musical features of different time scales, like repeats and notes of various lengths.

We have included a thorough evaluation of this improved model, employing a supercomputer and user feedback from our application in the gaming industry operation with a business company.
The benchmarks show that our improved model outperformed the state-of-the-art model known as the Dance Dance Convolution; DDC \cite{Donahue:2017}.
From the business feedback (\cref{sec:feedback}), we learned that the model is capable of replacing the manual labor of generating the first drafts (this is practical, since game artists have relied on manager-level decisions to fine-tune their first drafts anyway).
Our model, named \emph{GenéLive!}, reducing the business cost by half, will stay in business operation for the foreseeable future.

\begin{figure}[tb]
    \centering
    \includegraphics[width=\linewidth]{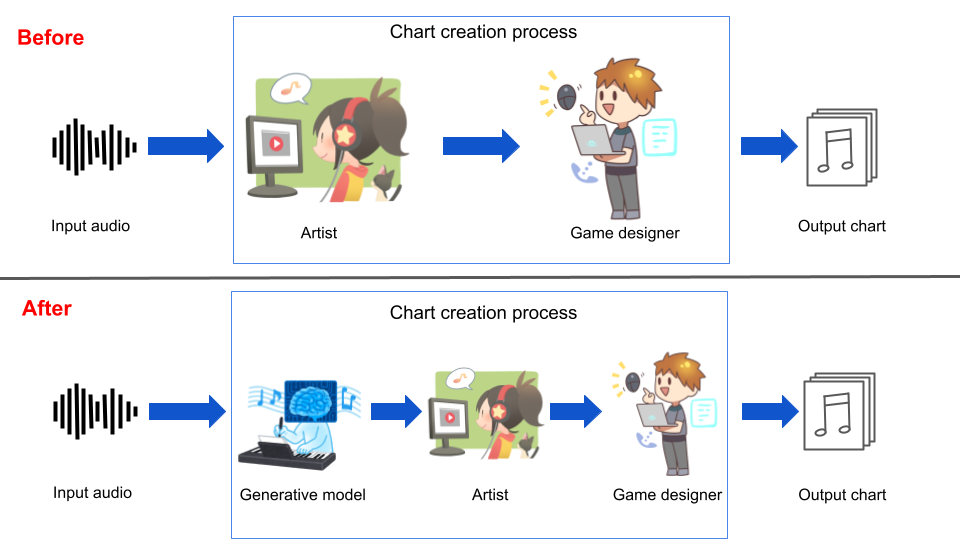}
    \caption{KLab's chart generation workflow before and after this work.}
    \label{fig:workflow}
    \includegraphics[width=\linewidth,clip,trim=120 0 200 0]{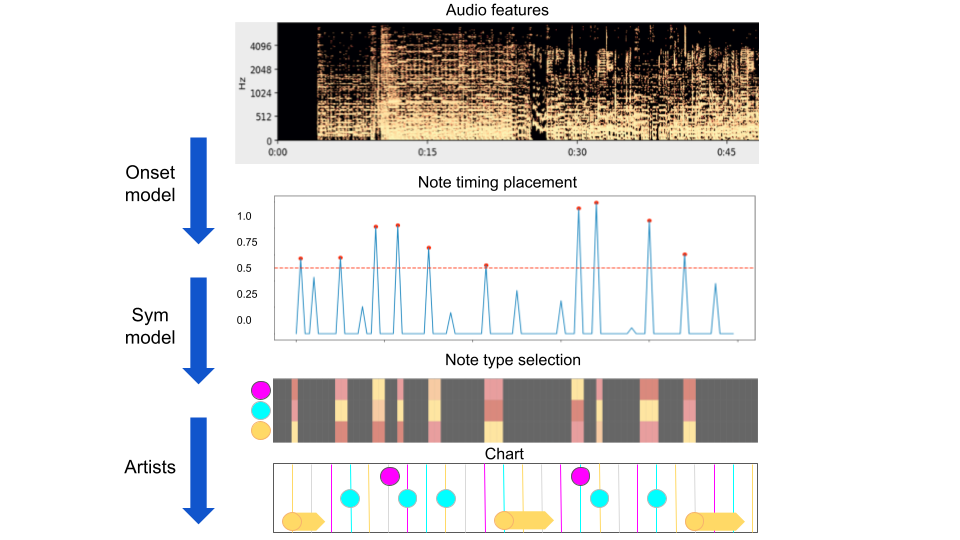}
    \caption{Data processing in the generative model. The present article focuses on the onset model.}
    \label{fig:chart_example}
\end{figure}

To conduct the application study, we worked with the game artists team in KLab Inc., a Japan-based video game developer (which employs also some of the authors).
The company has by today operated three rhythm action game titles online.
A particularly successful title, ``Love Live!~School Idol Festival All Stars,'' or simply ``Love Live!~All Stars'' (\cref{fig:game_screen}), has been released in 6 languages and been played worldwide while under KLab's operation, acquiring more than 10 million users.
Today, we see a wide range of competitive games with comparable impacts, which makes this work relevant to a larger audience.

Specifically, our target task is the generation of \emph{charts}, which instruct the player to tap or flick buttons at specified moments, the defining challenge of rhythm action games.
These buttons are known as \emph{notes} as they fly through the screen forming a spatial pattern resembling a musical score.
The audio record playing in the background is commonly referred to as a \emph{song}.

A chart generation model consists of two submodels: the \emph{onset}, which generates the timing of a note, and the \emph{sym}, which decides the user action type (like a tap or flick).
As deciding the timing is the bottleneck of KLab's workflow (detailed in \cref{sec:workflow}), this article focuses on presenting our onset model.

The present work has the following contributions:
\begin{itemize}
    \item We propose a deep generative model for chart creation, which achieved its business-quality performance by improving the state-of-the-art model DDC \citep{Donahue:2017} by incorporating two novel techniques: the beat guide and multi-scale conv-stack.

    \item Each of our improvements enhances the performance for all difficulty modes in multiple game titles. The improvements were effective particularly for easier difficulty modes, overcoming a commonly known weak point of the DDC.

    \item Incorporated into the workflow of KLab's rhythm action titles, our model halved the chart delivery time. The workflow is usable to rhythm actions in general -- the results verified the versatility also for open datasets from third parties, \emph{Stepmania}.

    \item Our PyTorch-based source code and the trained models, found after extensive hyperparameter tuning (over 80,000 GPU hours of Tesla P100 on a supercomputer), are publicly available.\footnote{\url{https://github.com/KLab/AAAI-23.6040}}
\end{itemize}

\section{Problem Definition}\label{sec:problem}

\subsection{Musical Structure}\label{sec:musicalStructure}
Most songs used in rhythm action games have a typical musical structure that is composed of temporally hierarchical performance patterns.
The percussion keeps a steady beat, creating a rhythmic pattern in a bar.
A series of such bars, together with phrases of melody instruments or vocals, forms musical sections such as an intro, verse, bridge, chorus, and outro.
For example, the song \emph{Sweet Eyes}\footnote{The song is available on \url{https://youtu.be/MpAUJ36fq3g}, and its portion from intro to first chorus (time range 0:16--2:07) is played in our rhythm game.} in ``Love Live!~All Stars'' has 60 bars that organize the following musical structure: an 8-bar intro, 16-bar verse, 10-bar bridge, 18-bar chorus, and 8-bar outro, each of which consists of repetitions of 1-bar to 4-bar phrases.

Artists at KLab confirm that this sort of musical structure is common in almost all songs in their rhythm action games and also a key feature of their chart creation.
More specifically, the artists tend to put identical note patterns on the above-mentioned repetitions of a phrase.
See \cref{sec:chart_full} for more details on the musical structure of Sweet Eyes and how the artists put notes on such a structure.

To learn the temporal patterns with our generative model, we would thus be required to consider multiple time scales. This was, in fact, absent in the network design of the DDC.
To see how we designed our model to capture these features, see the explanations of the beat guide (\cref{sec:beat-information-consideration}) and multi-scale conv-stack (\cref{sec:conv-stack}).

\subsection{Game Difficulties}
A song will be assigned charts of various difficulty modes ranging from \emph{Beginner}, \emph{Intermediate}, \emph{Advanced}, and \emph{Expert} to \emph{Challenge}, in increasing order.
In our preliminary experiments, the Dance Dance Convolution (DDC) \citep{Donahue:2017} generated charts for higher difficulty game modes at a human-competitive quality. (Find more related work in \cref{sec:related-work}.)
However, the generation of low-difficulty charts had room for improvement (as \citet{Donahue:2017} themselves pointed out).
As our primary target was easier modes, this was a significant challenge.

\subsection{Challenges in Business Application}\label{sec:workflow}
Leading rhythm action titles today tend to take the form of one piece of a large entertainment franchise, like KLab's ``Love Live!~All Stars'' does of ``Love Live!.'' 
The company's role is to operate the mobile app, while songs are delivered by other participants of the franchise.
After the first release of the app, KLab continued to contribute by offering new playable songs.
This is why a significant cost for the company's business is posed by chart generation.

The company's workflow (\cref{fig:workflow}) does \emph{not} demand a \emph{fully}-automated chart generation, since KLab's artists need to experiment with different variations of candidate products, employing their professional skills -- this is a high-level decision critical for the success of the franchise.
We thus focus on semi-automation, which was to generate the \emph{first drafts} of the charts (see \cref{fig:chart_example}) so the artists can be freed from this low-skill labor. 

To create a chart, artists repeatedly listen to the whole of a song to understand its musical structure as set by business partners.
During this process, they ponder how to place hundreds of notes to be tapped by the player, to eventually craft the chart through trial and error.
This first draft does not require too much expert skill, although it had been causing as much as half the cost in the workflow before our model was in operation.

The charts are then modified so that the actions are connected with emotions, like imitating the dance motion rendered in the background or flicking to a specific direction relating to the lyrics.
It may be revised further to enhance the gameplay experience with more focused consideration of the overall game design.

In essence, the first draft of the chart generation is crafted only from the input audio, while the enhancements are applied from information harder to compile into numerical data.
We thus target the auto-generation of the first drafts.

\section{Datasets}\label{sec:datasets}
\subsection{Datasets Acquired}\label{sec:data-acquisition}
We acquired songs and charts used in ``Love Live!~School Idol Festival All Stars'' (in short ``Love Live!~All Stars'') and ``Utano Princesama Shining Live'' (``Utapri'') operated by KLab.
Both the songs and charts are provided by multiple artists.
In addition, we use openly accessible songs and charts from ``Fraxtile'' and ``In the groove'' in the open source game ``Stepmania,'' which were used also in the prior work \citep{Donahue:2017}.
The number of songs were 163, 140, 90, and 133 for ``Love Live!~All Stars,'' ``Utapri,'' ``Fraxtile,'' and ``In the groove,'' respectively.
There were typically 4 game difficulties for ``Love Live!~All Stars'' and 5 for the rest, each difficulty contributing to one chart.
See \cref{sec:characteristics} for details on the datasets.

\subsection{Data Augmentation}
We augmented the audio of each song in the datasets.
The audio was first converted to a Mel spectrogram, which is a 2D array of time-frequency bins.
The spectrogram was then augmented via a series of transformations adopted from \citep{Park_2019}, resulting in an augmented Mel spectrogram, which is an input to generative models.

We applied the following transformations in the presented order (see \cref{sec:dataAugmentationDetails} for details):
the \emph{frequency shift} shifts all frequency bins by a random amount;
\emph{frequency mask} fills some frequency bins with the mean value;
\emph{time mask} fills some time bins with mean value;
\emph{high frequency mask} also fills some frequency bins but such frequencies must be above a predetermined threshold;
\emph{frequency flip} reverses the order of frequency bins;
\emph{white noise} adds a Gaussian noise to all time and frequency bins;
\emph{time stretch} stretches all time bins.

The onset labels, which specify the existence of notes in the chart, were augmented by fluctuating the labels \citep{Liang:2019}.
We also augment the beat information as explained in \cref{sec:dataAugmentationBeat}.

\section{GenéLive!}\label{sec:proposed-method}

\subsection{Audio Feature Extraction}\label{sec:audio-feature}
Following \citet{Donahue:2017}, our model uses the Short-Time Fourier Transform (STFT) and Mel spectrogram of the audio.
The STFT allows the model to capture features in the frequency domain.
The window length and stride of STFT were both set to be 32 ms.
The audio is sliced into chunks of 20 seconds.

The Mel spectrogram can capture perceptually relevant information in the audio data, and is a standard treatment in speech processing.
It is also used in the DDC \citep{Donahue:2017}.
Following Hawthorne et al.,~\citet{hawthorne:2017}, we set the number of the Mel bands to 229.
We set the lowest frequency to 0 kHz and the highest to 16 kHz (0 and 3575 in the Mel scale).
Accordingly, 229 evenly distributed triangular filters in the Mel scale are applied.
We denote a Mel spectrogram by $S(t,f)\in\mathbb R$, where $t=1,\dots,T$ denotes the $t$th time bin, and $f=1,\dots,F$ denotes the $f$th frequency bin.
We used $T = 20{,}000 / 32 = 625$ and $F=229$ as mentioned above.

\subsection{Base Model}
\begin{figure}[ht]
    \centering
    \includegraphics[width=\linewidth]{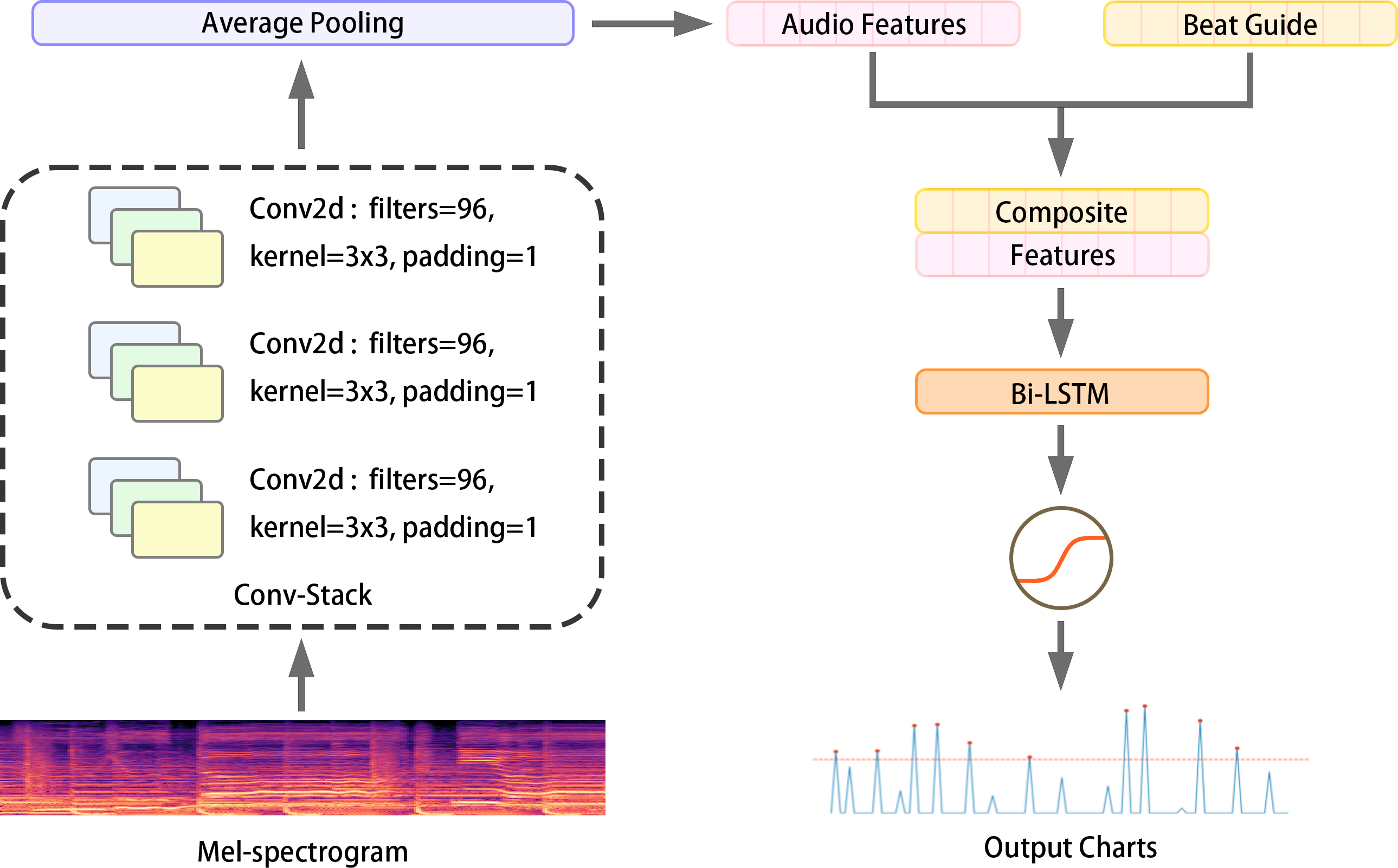}
    \caption{Overall architecture of our network.}
    \label{fig:base-model}
\end{figure}

As shown in \cref{fig:base-model}, our base model follows the DDC \citep{Donahue:2017}.
The Mel spectrogram $S(t,f)\in\mathbb R$ is processed through the CNN layers to extract audio features $A(t,f)\in\mathbb R$.
The audio features are concatenated with the game difficulty flag $D(t)=$ const.\ of 10 (Beginner), 20 (Intermediate), \dots, 50 (Challenge) and the beat guide $G(t)\in\{0,1,2\}$.
These two are fed to the BiLSTM layers \citep{Graves:2005} to generate the chart $C(t)\in [0,1]$.\footnote{We employed BiLSTM based on our preliminary experiments. More recent architectures, including Transformer-XL \citep{Dai2019}, performed worse than BiLSTM for our task.}
Our improvements are explained in sections \ref{sec:beat-information-consideration} and \ref{sec:conv-stack}.
Find more details of the model architecture and the corresponding parameters in \cref{sec:architecture}.

\paragraph{Convolution Stack}
The main task for the convolution stack (or \emph{conv-stack}) is to extract features from the Mel spectrogram using the CNN layers.
The conv-stack comprises a standard CNN layer with batch normalization, a max-pooling layer, and a dropout layer.
The activation function is ReLU. Finally, to regularize the output, we use an average-pooling layer.

\subsection{Beat Guide}\label{sec:beat-information-consideration}
\begin{figure}[ht]
    \centering
    \includegraphics[width=\linewidth,clip,trim=0 120 0 120]{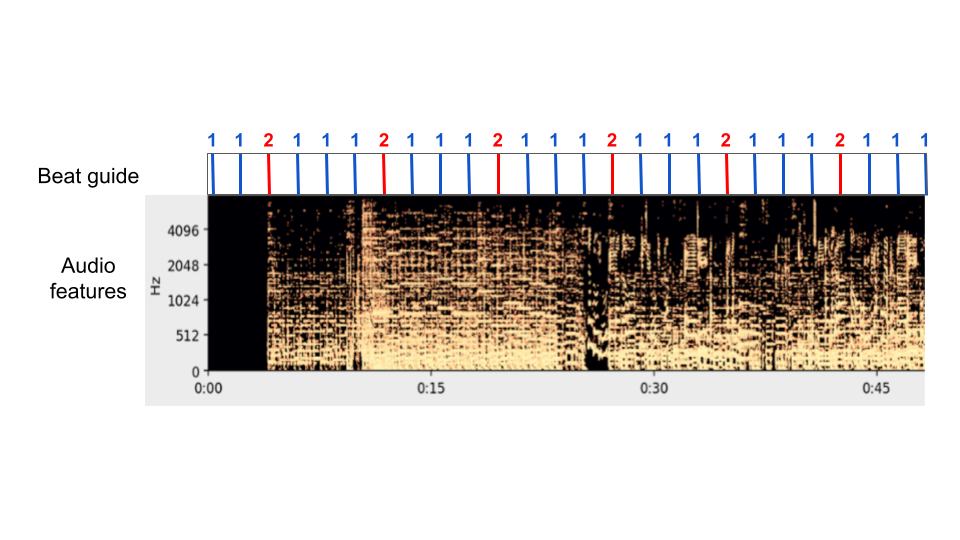}
    \caption{Beat guide in 4/4 time signature.}
    \label{fig:beat-guide}
\end{figure}

Although it had been rare to consider the positions of beats in the model, the beat is indeed crucial to the generation of the charts, as it is used by artists to evoke emotions.
The beat guide is a trinary array whose length is the same as the number of time frames of the input audio.
The first beat of each bar is indicated by 2, the other beats by 1, and non-beat frames by 0 (\cref{fig:beat-guide}).
Each element indicates the existence of a beat at that frame.
It is calculated from the BPM and time signature in the song metadata.
The beat guide is fed as an input to the BiLSTM layers. 

\subsubsection{Note Timings}\label{sec:timing}
\begin{figure}[hbt]
    \centering
    \includegraphics[width=0.75\linewidth]{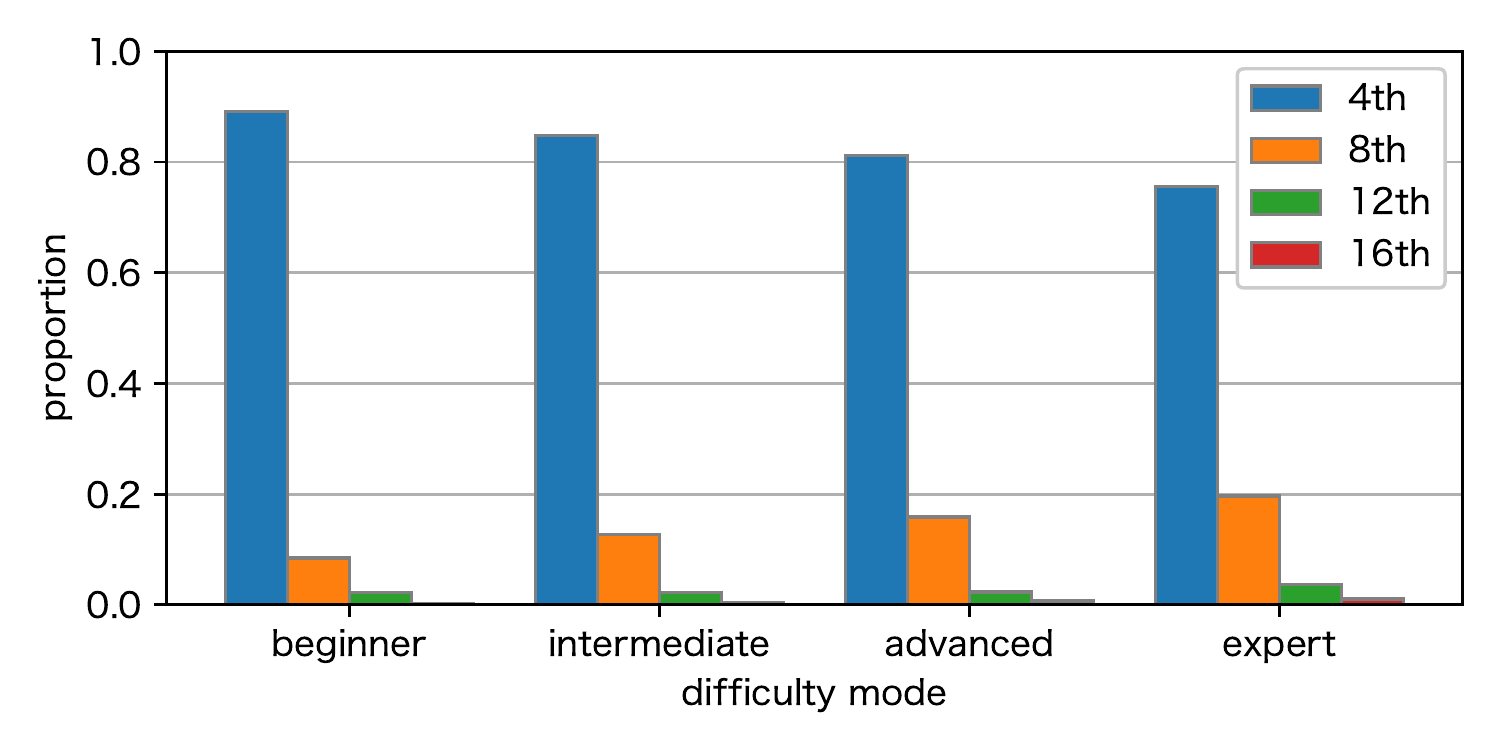}
    \caption{Note timings in ``Love Live!~All Stars.''}
    \label{fig:notes_stats}
\end{figure}

\Cref{fig:notes_stats} shows how frequently each note timing appears in KLab's charts.
The 4th note accounts for 70--90\% of a chart, and the 8th takes up 10--20\%; the 12th and 16th are marginal.
This fact supports the effectiveness of the beat guide, as it provides hints for placing 4th notes.
It also hints that the multi-scale conv-stack's temporal max-pooling layers would be able to extract temporal dependencies of the 4th and 8th note scales.

\subsubsection{Data Augmentation for Beat Guide} \label{sec:dataAugmentationBeat}
Since the proposed model requires a beat guide as an extra input accompanied with a Mel spectrogram, it is also augmented.
\emph{Beat mask} drops beats in the section with given probability.
The augmented guide is
\begin{equation}
    G(t) = \delta_t G_0(t),
\end{equation}
where $G_0(t)\in\{0,1,2\}$ is the original beat guide at time step $t=1,\dots,T$, and $\delta_t \sim \mathcal B(1,p)$ is a random number drawn for each $t$ from the binomial distribution with $p$ being the probability of dropping a beat.
The value of $p$ was optimized to 0.123 by random search in the range $[0.1, 0.3]$.
Finally, our model uses ($S$, $G$) as an input, where $S$ is an augmented Mel spectrogram defined in \cref{eq:augmentation} in appendix.

\subsection{Multi-Scale Conv-Stack}\label{sec:conv-stack}
\begin{figure}[ht]
    \centering
    \begin{minipage}[b]{0.25\linewidth}
        \centering
        \includegraphics[width=\linewidth]{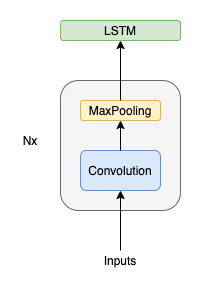}
        \subcaption{DDC}
        \label{fig:ddcconv}
    \end{minipage}%
    \begin{minipage}[b]{0.75\linewidth}
        \centering
        \includegraphics[width=\linewidth,clip,trim=40 560 60 0]{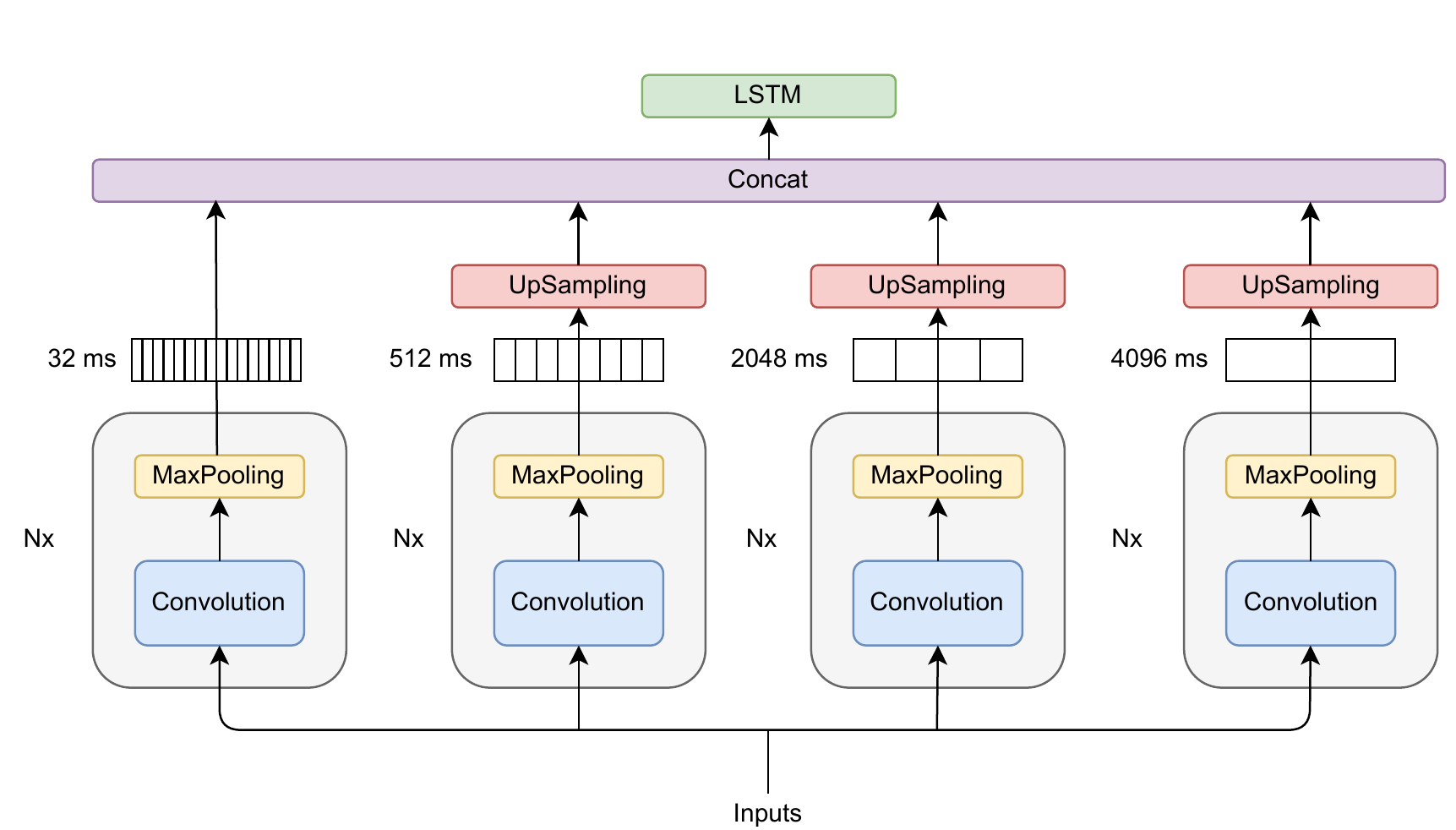}
        \subcaption{GenéLive!}\label{fig:klabconv}
    \end{minipage}
    \caption{Conv-stack architectures, previous vs.~present.}
    \label{fig:conv}
\end{figure}

One key difference between the DDC and the present model is the structure of the conv-stack.
In the model used in DDC, the convolution layers are applied repeatedly to the input of Mel spectrogram, whereas the max-pooling reduces the matrix size only along the frequency axis and not time (\cref{fig:ddcconv}).

The present model uses four conv-stacks with different temporal resolutions.
The stack with the highest resolution (stack 1) does not perform max-pooling along the temporal dimension.
The process is the same as the conv-stack of the DDC.
In stacks 2, 3, and 4, max-pooling is performed along the time dimension, and the length is reduced to $1/16$, $1/64$, and $1/128$, respectively.
Finally, up-sampling is applied to stacks 2, 3, and 4, and the four matrices, which have the same length in the temporal dimension, are concatenated (\cref{fig:klabconv}).
By doing so, we expect our model to extract not only short-term features (e.g., the attack of the percussion) but also long-term features (e.g., rhythm patterns and melodic phrases). 

Note that unlike generic 2D multi-scale convolutions such as GoogLeNet \citep{Szegedy2015}, our temporal max-pooling does multi-scale pattern extraction explicitly and only along the time axis. Existing networks can be distracted by multi-scale patterns in frequency arisen by instruments such as piano or trumpet. More important, as agreed with our artists, is multi-scale temporal patterns.

\begin{figure}[ht]
    \centering
    \includegraphics[width=\linewidth]{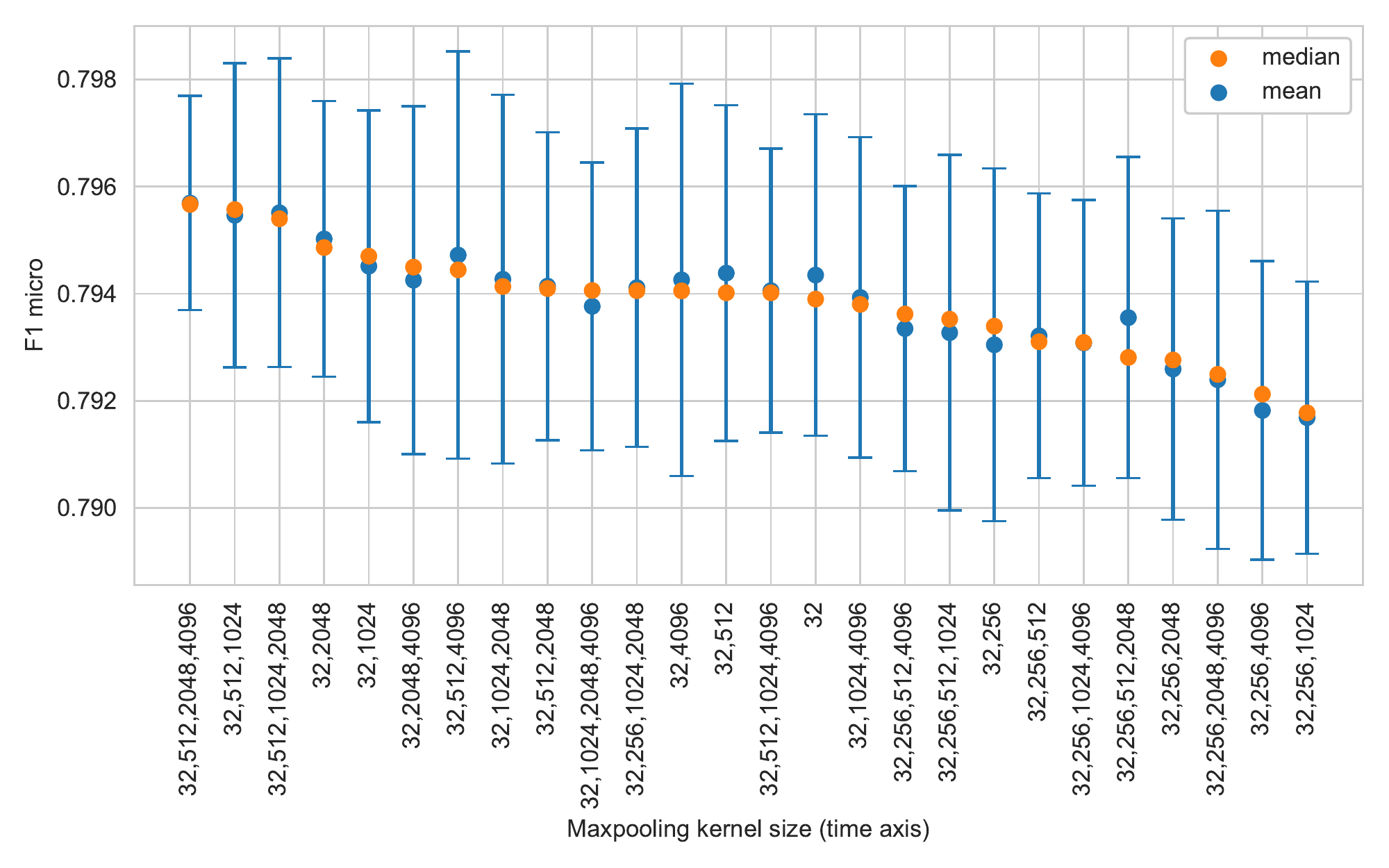}
    \caption{
        Experimenting with multi-scale temporal analysis.
        The results are sorted for the median of the F$_1$-score$^{\textrm m}$ in descending order along the horizontal axis. The error bar shows $1\sigma$ of results.
        The baseline is a single max-pooling of kernel size 32 ms (placed in the middle of the horizontal axis) that was employed in the DDC model. There is a statistical significance in our multi-scale model (32, 512, 2048, 4096) improving the baseline (32) (Wilcoxon rank sum test, $W=1703.5$, $p<0.01$).
    }
    \label{fig:convstack-combination}
\end{figure}

The results of taking a combination of different conv-stacks are shown in \cref{fig:convstack-combination}.

\subsubsection{Effectiveness of Multi-Scale Analysis}
\Cref{fig:convstack-combination} shows the effectiveness of multi-scale conv-stacks with different kernel sizes for max-pooling.
The size 32 ms is the baseline chosen also by the DDC (without multi-scaling).

For candidates of max-pooling kernel size, we choose lengths with regular intervals in logarithmic scale to be musically meaningful length: 256 ms, 512 ms, 1024 ms, 2048 ms, and 4096 ms, each of which corresponds to the 8th note, 4th note, 2nd note, one bar, and two bars at 120 beats per minute (BPM) in 4/4 time signature.\footnote{The BPM 120 is the rough average in our data set (71 min / 230 max). The performance is insensitive enough to this factor (\cref{fig:max-pooling} in \cref{sec:maxpooling-kernel-size}).}

We can see that the 8th note (256 ms) typically worsens the learning compared to 32 ms.
In this experiment combining 4 scales at maximum, the best one combines 32 ms, 512 ms (4th note), 2048 ms (one bar), and 4096 ms (two bars).

\section{Experiments}\label{sec:experiments}
For evaluation, we conducted a few experiments, while the business feedback is found in \cref{sec:feedback}.
We took the ablation approach for evaluation, in which the \emph{GenéLive!} model with all the presented methods applied was compared against models lacking some single component each.

As the characteristics of charts differ between different game titles, we conducted experiments separately for each game title.
To spare pages, the text focuses on the results of the ``Love Live!~All Stars'' dataset.
With other datasets including another private dataset for ``Utapri'' in KLab and open datasets for Stepmania, the model exhibits similar results, confirming the genericity of the performance (See \cref{sec:results-detail}).

\subsection{Training Methodology}\label{sec:training-methodology}
We used a supercomputer to train the model with a vast range of hyperparameter configurations.
In each training, we used the \emph{BCE} as the loss function.
Model parameters were updated using the \emph{Adam} optimizer \citep{kingma2014method}.
The cosine annealing scheduler \citep{Ilya:2016} tuned the learning rate for better convergence.
During the training, the dropout strategy was employed in both the fully connected layer and BiLSTM layer.

\subsubsection{Tuning Hyperparameters}\label{sec:hyperparameters}
The supercomputer let us conduct a grid search to determine the optimal combination of the following hyperparameters:
the learning rate, $\eta_\mathrm{min}$ in the \emph{cosine annealing scheduler} \citep{Ilya:2016}, the choice of conv-stack, the width and scale of \emph{fuzzy label} \citep{Liang:2019}, the dropout rate in the linear layer, the dropout rate in the RNN layer, the number of RNN layers, and the weighting factor in BCE loss.

To compare two sets of hyperparameters, we looked at the median of F$_1$-score$^{\textrm m}$ (as explained in \cref{sec:metrics}) for the validation dataset.

To see hyperparameter values used in the experiment and candidates of grid-search, see \cref{tab:hyperparameters} in \cref{sec:hyperparametersAppendix}.

\subsubsection{Computing environment}
We conducted the experiments using our supercomputer's NVIDIA Tesla P100 GPUs. 
We employed 64 of these GPUs to do the grid search in parallel.
The implementation is based on \verb|pytorch|, and to pre-process audio data \verb|librosa| was employed.
See \texttt{conda.yaml} file in the supplementary material for the complete list of software dependencies.

\paragraph{Data splitting}
The dataset was split into $8:1:1$ for training, evaluation, and testing sets with holdout employed.
The dataset was first split into a few subgroups with similar BPM.
The stratified K-folds cross-validator was then used to re-split each dataset into three.

Our model has no structural restriction on the time length of the input.
In practice, however, a whole song cannot be supplied due to memory capacity constraints and the difference in the duration of songs.
Thus, the audio, as well as the target, are cut into short chunks.
Since the chunk length has room for exploration, we treated it as  hyperparameters.

\subsubsection{Metrics}\label{sec:metrics}
Following \citet{Donahue:2017}, we use the F$_1$-score as the evaluation metric.

Firstly, the F$_1$-score is calculated by averaging the results for each chart.
We denoted the metric as F$_1$-score$^{\textrm c}$.
Charts with different difficulty modes for the same song are considered distinct.
Secondly, the score, F$_1$-score$^{\textrm m}$, is calculated by micro averaging.
We considered the predicted note to be true positive if the note is placed within $\pm 50$ ms around the ground truth. This is small enough for the shortest note (16th, which means 128 ms in case of 120 BPM) in our datasets.

We calculated the F$_1$-score for the test datasets using the median out of 10 experiments.
For each of the 10, evaluation metrics were calculated for every step.
the value with the highest performance for each evaluation metric was adopted as the performance of the model.

\subsection{Results}\label{sec:results}
Although a detailed discussion is to be found in \cref{sec:discussion},
we can see that our model consistently outperformed the state-of-the-art model \citep{Donahue:2017} as shown in \cref{tab:result}:
\begin{table}[ht]
    \centering
    \scriptsize
    \caption{Chart generation quality of the present model (GenéLive!) and the state-of-the-art (DDC) over 10 trials.
    Since in 10 trials the worst F$_1$-score of the proposed model is better than the best one of DDC in every difficulty mode, the one-sided Wilcoxon rank-sum test results in the possible smallest (identical) $p$-value, indicating that the proposed method statistically significantly outperforms over DDC.}
    \label{tab:result}
    \begin{tabular}{c | c c c}
        \toprule
        & \multicolumn{2}{c}{F$_1$-score$^{\textrm m}$ (mean $\pm$ SD)} & $p$-value\\
        Difficulty & GenéLive! & DDC & \\
        \midrule
        Beginner     & $0.8664 \pm 0.0036$ & $0.7839 \pm 0.0081$ & $7.85 \times 10^{-5}$\\
        Intermediate & $0.7950 \pm 0.0051$ & $0.7457 \pm 0.0039$ & $7.85 \times 10^{-5}$\\
        Advanced     & $0.7875 \pm 0.0045$ & $0.7491 \pm 0.0044$ & $7.85 \times 10^{-5}$\\
        Expert       & $0.7955 \pm 0.0038$ & $0.7603 \pm 0.0059$ & $7.85 \times 10^{-5}$\\
        all          & $0.8019 \pm 0.0026$ & $0.7514 \pm 0.0046$ & $7.85 \times 10^{-5}$\\
        \bottomrule
    \end{tabular}
\end{table}

The F$_1$-score is explained in \cref{sec:training-methodology}.
Although the metric is calculated for separate difficulties, the model was trained by being fed charts from all difficulties.

\begin{figure}[ht]
    \centering
    \includegraphics[width=\linewidth]{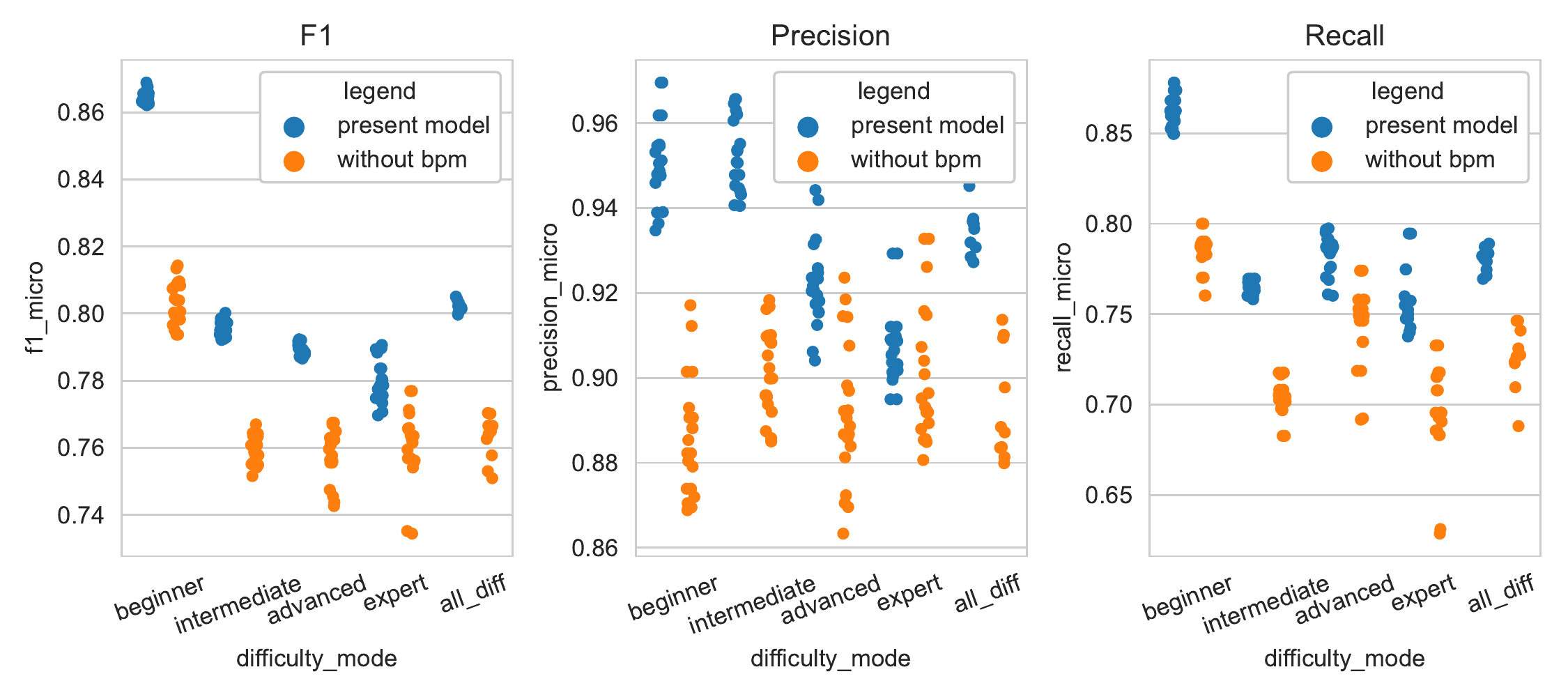}
    \caption{The beat guide enhanced the performance.}
    \label{fig:beat-ablation}
\end{figure}

While we have already seen the results indicating the effectiveness of the multi-scale conv-stack in \cref{fig:convstack-combination}, the same for the beat guide is shown in \cref{fig:beat-ablation}.

\begin{figure}[ht]
    \centering
    \includegraphics[width=\linewidth]{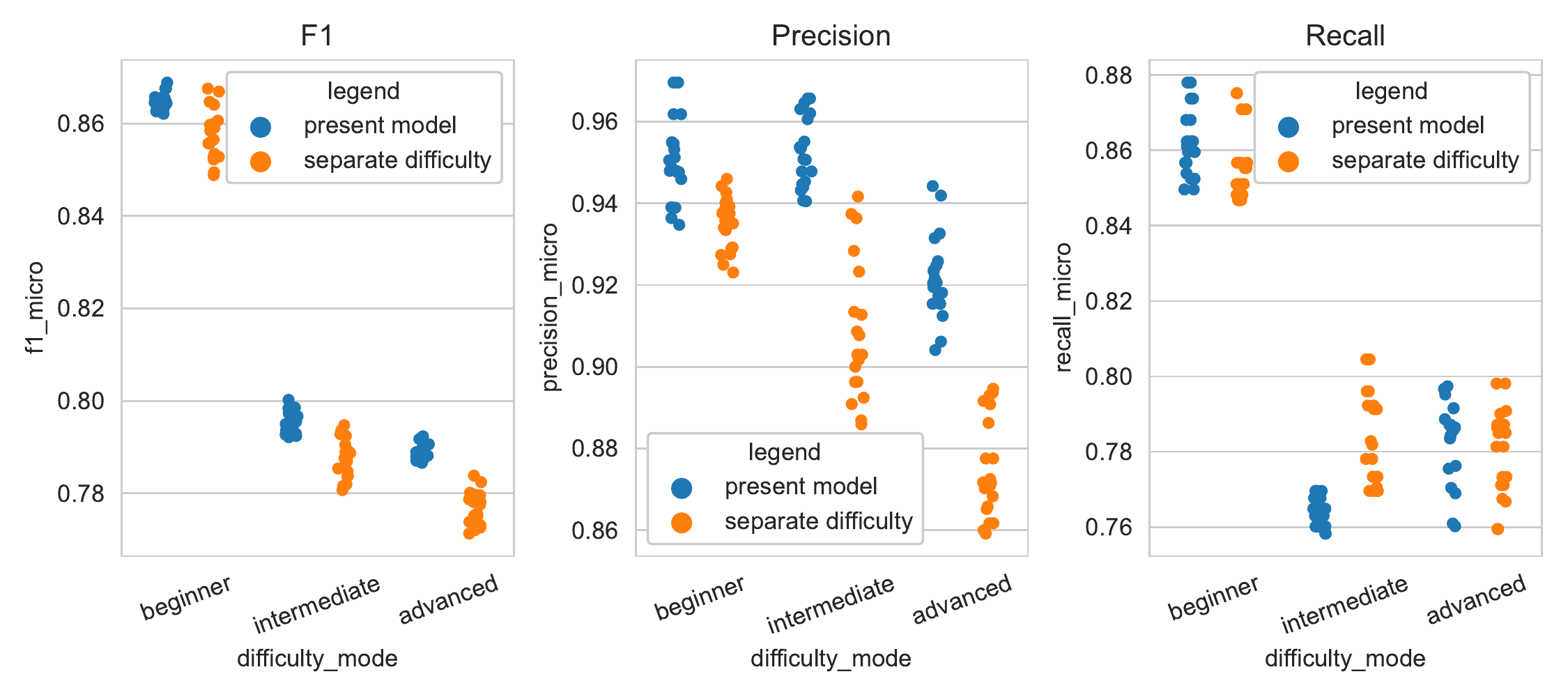}
    \caption{Performance with training using datasets from all difficulties vs a single difficulty.}
    \label{fig:difficulty-ablation}
\end{figure}

Our decision to train a single model using charts from all difficulty modes is justified by the results in \cref{fig:difficulty-ablation}, which is a comparison against training different models for different difficulty modes.

\subsection{Discussion} \label{sec:discussion}

\subsubsection{Beat Guide}
\Cref{fig:beat-ablation} shows that removing the beat guide significantly degraded the performance of the chart prediction task.
Compared with the other two experiments, the beat guide is the most significant element in the present model.

The model's CNN seems to value locations where the sound is loud. If the beat guide is absent they tend to fail at other rhythm patterns like a repeating sequence with moderate volume.

Comparing the results for different difficulty modes, we can see that the effect of adding the beat guide to the input was larger for the easier difficulty modes.
In general, charts of lower difficulty modes tended to consist of rhythms easier to capture.
Keeping the beat of the song is one of the simplest rhythms, so in charts of lower difficulty modes, the note is often placed at the beat positions as shown in \cref{fig:notes_stats}.
In turn, the easier difficulty mode of a chart, the more emphasis is placed on periodical rhythms, which CNN-based models are not good at, than the sound played in the music, and this may be the reason why the performance of the note generation task with easier difficulty modes is lower in previous studies such as \citep{Donahue:2017}.
Our experiments show that adding the beat guide to the input is a key to overcoming this weak point.

\subsubsection{Multi-Scale Conv-Stack}
\Cref{fig:convstack-combination} shows that the combination of multiple convolution stacks with appropriate max-pooling kernel size improves the performance, compared to the original conv-stack (denoted by 32 ms in the figure).
This is the effect of the model being able to ``look at'' the time direction not only in the BiLSTM layer but also in the CNN layer by introducing time-axis max-pooling.
The best combination of max-pooling kernel sizes is (32 ms, 512 ms, 2048 ms, 4096 ms).
Specifically, compared to the original conv-stack, the stack 2 of our multi-scale conv-stack is able to take into account 16 times coarser information in the time direction (as discussed in \cref{sec:conv-stack}).
Since the time resolution is 32 ms in our case (as explained in \cref{sec:audio-feature}), the second stack can consider 512 ms forward or backward, which corresponds to the length of a 4th note at 120 BPM.
Similarly, our stack 3 sees 64 times coarser information, which amounts to one musical bar length at 120 BPM in 4/4 time signature, and our stack 4 sees 128 times coarser information, which amounts to two musical bar length.

\subsubsection{Training With All Difficulties}

Before this work, it had been understood that the DDC was poor at generating charts for easier difficulties \citep{Donahue:2017}, yet it had been unclear what kind of improvements can be effective.
That is to ask, is it better to let a \emph{single} model instance consume charts having multiple difficulties or, instead, \emph{multiple} instances consume them, where each instance is specialized for a certain difficulty mode?
On one hand, the similarity between charts from different difficulties for the same song could work as a hint. On the other hand, however, the network might be confused by the same song resulting in different charts.
Another question regarding the poor performance of the DDC on easier difficulties is whether the fundamentals of its design had a flaw in the first place.

\Cref{fig:difficulty-ablation} shows that when all difficulty modes were trained with a single model, the performance improved.
This result implies the effectiveness of the CNN for learning charts of varying difficulties.
This work is the first to reveal that learning different difficulties with a single model instance in fact outperforms the other.
The same experiments of ours also indicate that the DDC's approach, which we adapted, indeed is capable of predicting easy difficulties, although our model had received improvements.

What is the reason behind, though? In the next paragraph, we show our analysis that indicates a strong inclusion relation between charts for different difficulties but of the same song, which we suspect is the reason.

\subsubsection{Inclusion Relation Between Difficulties}\label{sec:inclusion}
We view a chart as a bit string representing the existence of a tap for frames in a quantized time.
We define the \emph{inclusion rate} as
\begin{equation}\label{eq:inclusion-rate}
    I(s,t) = \frac{\|s\otimes t\|}{\|s\|}
\end{equation}
for strings $s$ and $t$, where $\otimes$ is the bit-wise AND product, and $\|\cdot\|$ is the number of 1s in a string.
Inclusion rates (averaged over songs) in \cref{tab:inclusion-factor-lovelive} show that different difficulty modes actually share a large portion of notes.
Especially, more than 97\% of notes in an easier chart appears also in a harder one, indicating that training with easier charts contributes to the prediction of harder ones.
This pattern is also found in all datasets we targeted (see \cref{sec:characteristics}), and thus seems universal.

\begin{table}[ht]
    \small
    \centering
    \caption{Inclusion rates in ``Love Live!~All Starts.''
    }
    \label{tab:inclusion-factor-lovelive}
    \begin{tabular}{c|cccc}
        \toprule
        \diagbox[height=1.2\line,width=6em]{$t$}{$s$} & Beginner & Intermediate & Advanced & Expert \\ 
        \midrule
        Beginner & 100\% & 64.7\% & 49.8\% & 39.0\% \\ 
        Intermed.& 97.4\% & 100\% & 76.2\% & 59.4\% \\ 
        Advanced & 97.4 \% & 98.8 \% & 100\% & 73.5\% \\ 
        Expert &   98.6 \% & 98.2\% & 98.6\% & 100\% \\ 
        \bottomrule
    \end{tabular}
\end{table}

\section{Business Feedback}\label{sec:feedback}
Since the first deployment of this chart generation system (July 2020), the artist team has used the system to create charts for all 110 songs released in ``Love Live!~All Stars'' (82 songs had been released before the period).
The present collaboration cut down the chart creation time for the artist team by half.
About 20 hours of work are saved per song, whereas it used to take about 40 hours per song.
For the Beginner and Intermediate difficulty modes, the charts generated by our model can be used with minor modifications.
The artist team uses our model also for the more difficult game modes.

\begin{figure}[htb]
    \centering
    \includegraphics[width=1.1\linewidth]{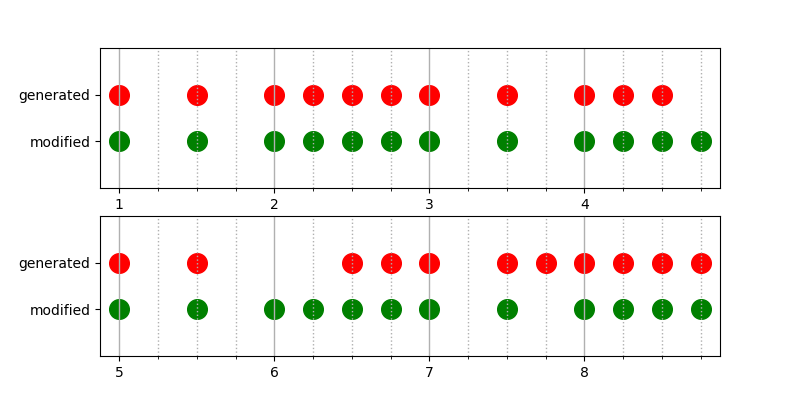}
    \caption{A generated chart for the Advanced difficulty mode and its manually-modified version (the first 8 bars).}
    \label{fig:chart}
\end{figure}

Artists employed the timing for notes generated by the model mostly without alteration.
\Cref{fig:chart} compares the first 8 bars of an automatically generated chart with the released version which was manually modified from the generated one.
Of the 22 auto-generated notes, 21 were accepted as they were, and only one was not used, while three notes were added.
Such a high-quality chart is mainly brought by our model's novel ability to recognize musical structures.
As you can listen to the song on YouTube\footnote{The song is available on \url{https://youtu.be/MpAUJ36fq3g}, and its portion from intro to first chorus (time range 0:16--2:07) is played in our rhythm game.}, this 8-bar intro repeats a 2-bar phrase where the phrase is altered every repetition.
Our model reflected such an altered repetition of a 2-bar pattern in the generated chart (\cref{fig:chart}).
This is true for the whole of the song, see \cref{fig:chart-full} in \cref{sec:chart_full}.

\section{Related Work}\label{sec:related-work}
Early chart generation tended to use a rule-based algorithm \citep{o2003dancing} or genetic algorithm \citep{nogaj2005genetic}.
The DDC by \citet{Donahue:2017} improved the quality by employing the deep neural network.
\citet{Lin:2019} used a multilayer feed-forward network to generate charts, synchronizing the notes with instrumental sounds.
The model by \citet{Tsujino:2018} alters the difficulty of an input chart -- this work instead generates the first draft chart (without an input chart). 

Onset detection based on neural networks is studied for speech recognition, where BiLSTMs \citep{eyben2010universal}, CNN approach \citep{schluter2014improved}, and multi-resolution feature representation are established.
\citet{schluter2014improved} demonstrated a CNN-based approach.
Likewise, music information retrieval has also seen benefits of neural networks \citep{humphrey2012rethinking,boulanger2013audio,ullrich2014boundary}.

In addition, various deep generative models are used to create game content and assets.
\emph{Procedural content generation via machine learning}, a model for generating game content using trained machine learning models, can generate a variety of game content such as items, maps, and rules \citep{summerville2018,guzdial2018explainable}.
\citet{Erin2009} proposed an automatic content generation that learns players' preferences based on their past play history and generates new graphical and game content during gameplay.
In addition, \citet{cerny2018press} proposed a framework generating tutorial text by discovering whether it is effective to win or lose a game.
\citet{tilson2019} investigated generating image assets for games using unsupervised learning such as GAN and VAE.

\citet{volz2018evolving} have generated various levels of Super Mario Bros.~using GAN.
\citet{park2019} used GAN to generate levels for a computer science educational game.

There is a method to synthesize high-quality, realistic full-body animations of 3D characters using deep learning \citep{YuDing2021}.
It is possible to express emotions and generate facial expressions and movements corresponding to the NPC's personality and profession.
A model has been developed to create a face mesh from a photograph of a face and generate a face model of a 3D character \citep{PeiLi2021}.

In addition, there are examples of successful use of deep learning to reduce man-hours in the game development process, such as the development of AI for mini-games in MMORPGs by applying AlphaZero \citep{Lei2021} and the improvement of game frame rates by applying deep super-resolution in the time direction \citep{Andrew2021}.

\section{Conclusions}\label{sec:conclusions}
We assisted in chart creation at KLab Inc.
by establishing a new deep generative model, while the model ended up being in fact versatile for more generic datasets.
The model successfully generated charts for easier difficulty modes, filling the quality gap between easier and harder game modes, which was a challenge that had been admitted by the authors of the state-of-the-art model DDC.
This was achieved by utilizing two techniques, (i) the beat guide and (ii) the multi-scale conv-stack.
KLab successfully cut the business cost by half.

\section*{Acknowledgements}
The authors would like to say \emph{arigatou} to every member of $\mu$'s, Aqours, Nijigaku, Liella!\ and all school idols, as well as Lovelivers (fans) over the world, whose energy of \emph{daisuki} always encouraged us and drove our research.
We are especially grateful to Lingjian Wang and Keita Yamamoto for their development and operation of our chart generation system, Chiharu Shineha, Yuko Okada and Takahiko Anbo for their insightful comments from the professional viewpoint of chart creation.
This work was supported by JSPS KAKENHI Grant Number 20K19809.
The computation was carried out using the computer resource offered
under the category of Intensively Promoted Projects by Research Institute for Information Technology, Kyushu University.

\bibliography{main}

\clearpage
\appendix
\section{Characteristics of Datasets} \label{sec:characteristics}

\Cref{tab:dataset-stats} summarizes the songs and charts in the datasets.

\begin{table}[ht]
    \centering
    \caption{Datasets. Stepmania F refers to Stepmania Fraxtil.
    Charts of the same song with different difficulty modes are counted as different charts.
    }
    \label{tab:dataset-stats}
    \begin{tabular}{c|ccc} 
        \toprule
        Datasets & \# songs & \# charts & \# difficulty modes \\ 
        \midrule
        Love Live! & 163 & 501 & 4 \\
        Utapri & 140 & 579 & 5 \\
        Stepmania F & 90 & 450 & 5 \\
        Stpmania ITG & 133 & 652 & 5 \\
        \bottomrule
    \end{tabular}
\end{table}

\Cref{tab:inclusion-factor-utapri,tab:inclusion-factor-stepmania} shows inclusion rates between charts of different difficulty modes for each dataset.
Looking at the upper-right triangle of the tables, we can see that charts created in our company have higher inclusion rates.

\begin{table}[ht]
    \centering
    \scriptsize
    \caption{Inclusion rates between difficulties of ``Utapri.''}
    \label{tab:inclusion-factor-utapri}
    \begin{tabular}{c|ccccc} 
        \toprule
        \diagbox[height=1.2\line,width=6em]{$s$}{$t$} & Beginner & Intermediate & Advanced & Expert & Challenge \\ 
        \midrule
        Beginner & 100\% & 91.0\% & 91.6\% & 92.4\% & 94.7\% \\ 
        Intermediate & 60.2\% & 100\% & 93.9\% & 94.3\% & 96.6\% \\ 
        Advanced & 37.8\% & 58.9\% & 100\% & 96.1\% & 96.5\% \\ 
        Expert & 25.9\% & 40.4\% & 65.8\% & 100\% & 96.7\% \\ 
        Challenge & 18.9\% & 30.2\% & 47.2\% & 69.7\% & 100\% \\ 
        \bottomrule
    \end{tabular}
\end{table}

\begin{table}[ht]
    \centering
    \scriptsize
    \caption{Inclusion rates between difficulties of ``Stepmania.''}
    \label{tab:inclusion-factor-stepmania}
    \begin{tabular}{c|ccccc} 
        \toprule
        \diagbox[height=1.2\line,width=6em]{$s$}{$t$} & Beginner & Easy & Medium & Hard & Challenge \\ 
        \midrule
        Beginner & 100\% & 91.2\% & 90.2\% & 88.8\% & 86.5\% \\ 
        Easy & 37.1\% & 100\% & 88.4\% & 86.3\% & 84.5\% \\ 
        Medium & 23.8\% & 57.2\% & 100\% & 87.1\% & 84.7\% \\ 
        Hard & 16.0\% & 37.9\% & 59.1\% & 100\% & 85.8\% \\ 
        Challenge & 11.0\% & 27.4\% & 42.5\% & 63.8\% & 100\% \\ 
        \bottomrule
    \end{tabular}
\end{table}

\section{Additional Experiments}\label{sec:results-detail}
In this section, we present additional experiments on one more performance metric, F$_1$-score$^{\textrm c}$, which was employed in the DDC's experiments \citep{Donahue:2017} accompanied with F$_1$-score$^{\textrm m}$, and three more datasets than ``Love Live!~All Stars,'' which are introduced in \cref{sec:datasets}; ``Utapri,''
``Stepmania Fraxtil,'' and ``Stepmania ITG\@.''
Note that ``Stepmania Fraxtil'' and ``Stepmania ITG'' are combined and then trained and evaluated.
They are denoted as ``Stepmania'' in tables below.

\Cref{tab:fscore-ddc} shows F$_1$-score$^{\textrm c}$ on the ``Love Live!~All Stars'' dataset.
Compared with \cref{tab:result}, we can see that F$_1$-score$^{\textrm m}$ and F$_1$-score$^{\textrm c}$ show a consistent performance.
\begin{table}[ht]
    \centering
    \scriptsize
    \caption{Chart generation quality of the present model (GenéLive!) and the state-of-the-art (DDC) in another performance metric, F$_1$-score$^{\textrm c}$, used in \citep{Donahue:2017}.}
    \label{tab:fscore-ddc}
    \begin{tabular}{c | c c | c}
        \toprule
        & \multicolumn{2}{c|}{F$_1$-score$^{\textrm c}$ (mean $\pm$ SD)} & $p$-value\\
        Difficulty & GenéLive!~& DDC \\
        \midrule
        Beginner     & $0.8654 \pm 0.0035$ & $0.7845 \pm 0.0080$ & $7.85 \times 10^{-5}$\\
        Intermediate & $0.7943 \pm 0.0049$ & $0.7471 \pm 0.0045$ & $7.85 \times 10^{-5}$\\
        Advanced     & $0.7870 \pm 0.0043$ & $0.7498 \pm 0.0042$ & $7.85 \times 10^{-5}$\\
        Expert       & $0.7972 \pm 0.0039$ & $0.7599 \pm 0.0058$ & $7.85 \times 10^{-5}$\\
        all          & $0.8104 \pm 0.0026$ & $0.7562 \pm 0.0049$ & $7.85 \times 10^{-5}$\\
        \bottomrule
    \end{tabular}
\end{table}

\Cref{tab:baseline} shows the performance of the present model, GenéLive!, for each dataset.

\begin{table}[ht]
\centering
\scriptsize
\caption{Performance of the present model (GenéLive!), mean $\pm$ SD over 10 trials.}
\label{tab:baseline}
\begin{tabular}{c c | c c}
\toprule
Dataset & Difficulty & F$_1$-score$^{\textrm m}$ & F$_1$-score$^{\textrm c}$ \\
\midrule
\multirow{5}{*}{Love Live!~All Stars}
 & Beginner & $0.8664 \pm 0.0036$ & $0.8654 \pm 0.0035$ \\
 & Intermediate & $0.7950 \pm 0.0051$ & $0.7943 \pm 0.0049$ \\
 & Advanced & $0.7875 \pm 0.0045$ & $0.7870 \pm 0.0043$ \\
 & Expert & $0.7955 \pm 0.0038$ & $0.7972 \pm 0.0039$ \\
 & all & $0.8019 \pm 0.0026$ & $0.8104 \pm 0.0026$ \\
\midrule
\multirow{6}{*}{Utapri}
 & Beginner & $0.6859 \pm 0.0083$ & $0.6858 \pm 0.0099$ \\
 & Intermediate & $0.7836 \pm 0.0037$ & $0.7802 \pm 0.0043$ \\
 & Advanced & $0.7678 \pm 0.0027$ & $0.7678 \pm 0.0038$ \\
 & Expert & $0.7979 \pm 0.0034$ & $0.7891 \pm 0.0035$ \\
 & Challenge & $0.8561 \pm 0.0037$ & $0.8561 \pm 0.0037$ \\
 & all & $0.7649 \pm 0.0024$ & $0.7443 \pm 0.0026$ \\
\midrule
\multirow{6}{*}{Stepmania}
 & Beginner & $0.7217 \pm 0.0153$ & $0.7093 \pm 0.0149$ \\
 & Intermediate & $0.7284 \pm 0.0037$ & $0.7239 \pm 0.0028$ \\
 & Advanced & $0.7428 \pm 0.0020$ & $0.7432 \pm 0.0016$ \\
 & Expert & $0.7349 \pm 0.0012$ & $0.7312 \pm 0.0018$ \\
 & Challenge & $0.7761 \pm 0.0031$ & $0.7731 \pm 0.0022$ \\
 & all & $0.7456 \pm 0.0023$ & $0.7303 \pm 0.0051$ \\
\bottomrule
\end{tabular}
\end{table}

\Cref{tab:beat} shows the performance of models trained without the beat guide.
Compared to the present model, the performance is significantly degraded in all datasets.
Furthermore, the degradation is more significant in easier difficulty modes.

\begin{table}[ht]
\centering
\scriptsize
\caption{Performance of GenéLive!\ without the beat guide, mean $\pm$ SD over 10 trials.}
\label{tab:beat}
\begin{tabular}{c c | c c c}
\toprule
Dataset & Difficulty & F$_1$-score$^{\textrm m}$ & F$_1$-score$^{\textrm c}$ \\
\midrule
\multirow{5}{*}{Love Live!~All Stars}
 & Beginner & $0.8034 \pm 0.0063$ & $0.8048 \pm 0.0065$ \\
 & Intermediate & $0.7594 \pm 0.0043$ & $0.7609 \pm 0.0043$ \\
 & Advanced & $0.7577 \pm 0.0077$ & $0.7595 \pm 0.0074$ \\
 & Expert & $0.7604 \pm 0.0110$ & $0.7619 \pm 0.0107$ \\
 & all & $0.7627 \pm 0.0067$ & $0.7694 \pm 0.0052$ \\
\midrule
\multirow{6}{*}{Utapri}
 & Beginner & $0.4520 \pm 0.0243$ & $0.4273 \pm 0.0252$ \\
 & Intermediate & $0.6560 \pm 0.0134$ & $0.6498 \pm 0.0132$ \\
 & Advanced & $0.6783 \pm 0.0092$ & $0.6777 \pm 0.0091$ \\
 & Expert & $0.7309 \pm 0.0099$ & $0.7237 \pm 0.0106$ \\
 & Challenge & $0.8218 \pm 0.0121$ & $0.8218 \pm 0.0121$ \\
 & all & $0.6739 \pm 0.0106$ & $0.6207 \pm 0.0137$ \\
\midrule
\multirow{6}{*}{Stepmania}
 & Beginner & $0.5944 \pm 0.0165$ & $0.5921 \pm 0.0165$ \\
 & Intermediate & $0.6874 \pm 0.0081$ & $0.6790 \pm 0.0079$ \\
 & Advanced & $0.7226 \pm 0.0046$ & $0.7220 \pm 0.0038$ \\
 & Expert & $0.7302 \pm 0.0046$ & $0.7265 \pm 0.0046$ \\
 & Challenge & $0.7768 \pm 0.0039$ & $0.7755 \pm 0.0040$ \\
 & all & $0.7290 \pm 0.0039$ & $0.6907 \pm 0.0043$ \\
\midrule
\end{tabular}
\end{table}

\Cref{tab:difficulty} shows the performance of models which is trained and evaluated separately for each difficulty mode.
In ``Utapri'' dataset, the performance is better when a single model is trained with charts of all difficulty modes.
However, ``Stepmania'' dataset didn't reproduce the result.
We are thinking that this is due to the lower inclusion rate of charts in ``Stepmania'' than that of charts created in our company, comparing \cref{tab:inclusion-factor-lovelive,tab:inclusion-factor-utapri,tab:inclusion-factor-stepmania}.

\begin{table}[ht]
\centering
\scriptsize
\caption{Performance of GenéLive!\ trained difficulty-wise, mean $\pm$ SD over 10 trials.}
\label{tab:difficulty}
\begin{tabular}{c c | c c c}
\toprule
Dataset & Difficulty & F$_1$-score$^{\textrm m}$ & F$_1$-score$^{\textrm c}$ \\
\midrule
\multirow{3}{*}{Love Live!~All Stars}
 & Beginner & $0.8577 \pm 0.0054$ & $0.8554 \pm 0.0057$ \\
 & Intermediate & $0.7877 \pm 0.0042$ & $0.7880 \pm 0.0043$ \\
 & Advanced & $0.7766 \pm 0.0036$ & $0.7766 \pm 0.0037$ \\
\midrule
\multirow{4}{*}{Utapri}
 & Beginner & $0.6411 \pm 0.0108$ & $0.6362 \pm 0.0146$ \\
 & Intermediate & $0.76763 \pm 0.0064$ & $0.7626 \pm 0.0075$ \\
 & Advanced & $0.7613 \pm 0.0040$ & $0.7622 \pm 0.0044$ \\
 & Expert & $0.7890 \pm 0.0031$ & $0.7800 \pm 0.0049$ \\
\midrule
\multirow{5}{*}{Stepmania}
 & Beginner & $0.7607 \pm 0.0022$ & $0.7513 \pm 0.0024$ \\
 & Intermediate & $0.7274 \pm 0.0039$ & $0.7228 \pm 0.0033$ \\
 & Advanced & $0.7477 \pm 0.0028$ & $0.7473 \pm 0.0026$ \\
 & Expert & $0.7303 \pm 0.0046$ & $0.7271 \pm 0.0046$ \\
 & Challenge & $0.7758 \pm 0.0041$ & $0.7711 \pm 0.0042$ \\
\bottomrule
\end{tabular}
\end{table}

\Cref{tab:convstack} shows the performance of models with the DDC's conv-stack in place of our multi-scale conv-stack.
We can see that there are certain performance improvements when the newer conv-stack is adopted, in all datasets.

\section{Model Architectures}\label{sec:architecture}
\Cref{tab:model_composition,tab:model_composition_two} show the conv-stack architecture of the DDC and our GenéLive!

\begin{table*}[t]
\centering
\scriptsize
\caption{Multi-scale conv-stack of GenéLive!}
\label{tab:model_composition}
\begin{tabular}{c | c | c | c | c}
\toprule
Block &  Conv stack 1 & Conv stack 2 & Conv stack 3 & Conv stack 4 \\
\midrule
\multirow{3}{*}{Convolution 1} & Kernel 3x3, Channels 48, Padding 1 & Kernel 3x3, Channels 48, Padding 1 & Kernel 3x3, Channels 48, Padding 1 & Kernel 3x3, Channels 48, Padding 1 \\
\cline{2-5}
& Batch Norm (48) & Batch Norm (48) & Batch Norm (48) & Batch Norm (48) \\
\cline{2-5}
& ReLU & ReLU & ReLU & ReLU \\
\midrule
\multirow{5}{*}{Convolution 2} 
 & Kernel 3x3, Channels 48, Padding 1  & Kernel 3x3, Channels 48, Padding 1  & Kernel 3x3, Channels 48, Padding 1 & Kernel 3x3, Channels 48, Padding 1 \\
\cline{2-5}
& Batch Norm (48) & Batch Norm (48) & Batch Norm (48) & Batch Norm (48) \\
 \cline{2-5}
 & ReLU  & ReLU  & ReLU & ReLU \\
 \cline{2-5}
 & MaxPool (1, 4) & MaxPool (4, 4) & MaxPool (4, 4) & MaxPool (4, 4) \\
\cline{2-5}
& Dropblock(0.25, 5, 0.25) & Dropblock(0.25, 5, 0.25) & Dropblock(0.25, 5, 0.25) & Dropblock(0.25, 5, 0.25) \\
\midrule
\multirow{5}{*}{Convolution 3} 
 & Kernel 3x3, Channels 96, Padding 1  & Kernel 3x3, Channels 96, Padding 1  & Kernel 3x3, Channels 96, Padding 1 & Kernel 3x3, Channels 96, Padding 1 \\
 \cline{2-5}
& Batch Norm (96) & Batch Norm (96) & Batch Norm (96) & Batch Norm (96) \\
 \cline{2-5}
 & ReLU  & ReLU  & ReLU & ReLU \\
 \cline{2-5}
 & MaxPool (1, 4) & MaxPool (2, 4) & MaxPool (2, 4) & MaxPool (2, 4) \\
 \cline{2-5}
& Dropblock (0.25, 3, 1) & Dropblock (0.25, 3, 1) & Dropblock (0.25, 3, 1) & Dropblock (0.25, 3, 1) \\
\midrule
\multirow{5}{*}{Convolution 4}
 & Kernel 3x3, Channels 384, Padding 1  & Kernel 3x3, Channels 192, Padding 1  & Kernel 3x3, Channels 192, Padding 1 & Kernel 3x3, Channels 192, Padding 1 \\
 \cline{2-5}
& Batch Norm (384) & Batch Norm (192) & Batch Norm (192) & Batch Norm (192) \\
 \cline{2-5}
 & ReLU  & ReLU  & ReLU & ReLU \\
 \cline{2-5}
 & MaxPool (1, 4) & MaxPool (8, 4) & MaxPool (32, 4) & MaxPool (64, 4) \\
\cline{2-5}
& AvgPool(1,3) & AvgPool(1,3) & AvgPool(1,3) & AvgPool(1,3) \\
 \midrule
\multirow{1}{*}{Up Sampling} 
& & Upsample (16, 1) & Upsample (64, 1) & Upsample (128, 1)\\
\midrule
\multirow{1}{*}{Functional 1} 
& \multicolumn{4}{c}{Dropout(0.5)} \\
\midrule
\multirow{2}{*}{BiLSTM} & 
\multicolumn{4}{c}{770 Features, 2 Layers, Bidirectional}\\
\cline{2-5}
& \multicolumn{4}{c}{Dropout(0.5)}\\
\midrule
\multirow{2}{*}{Functional 2} & \multicolumn{4}{c}{Linear 768, 1}\\
\cline{2-5}
& \multicolumn{4}{c}{Sigmoid} \\
\bottomrule
\end{tabular}
\end{table*}

\begin{table}[H]
    \centering
    \footnotesize
    \caption{The DDC's conv-stack.}
    \label{tab:model_composition_two}
    \begin{tabular}{c | c}
        \toprule
        Block &  Layers \& Parameters  \\
        \midrule
        \multirow{3}{*}{Convolution 1} & Kernel 3x3, Channels 48, Padding 1\\
        \cline{2-2}
        & Batch Norm (48) \\
        \cline{2-2}
        & ReLU \\
        \midrule
        \multirow{5}{*}{Convolution 2} 
         & Kernel 3x3, Channels 48, Padding 1 \\
        \cline{2-2}
        & Batch Norm (48) \\
         \cline{2-2}
         & ReLU \\
        \cline{2-2}
        & MaxPool (1,2) \\
        \cline{2-2}
        & Dropout (0.25) \\
        \midrule
        \multirow{5}{*}{Convolution 3} 
         & Kernel 3x3, Channels 96, Padding 1 \\
         \cline{2-2}
        & Batch Norm (96) \\
         \cline{2-2}
         & ReLU \\
        \cline{2-2}
        & MaxPool (1,2)\\
        \cline{2-2}
        & Dropout(0.25) \\
        \midrule
        \multirow{2}{*}{Functional 1} 
        & Linear 96*input//4, 48x16\\
        \cline{2-2}
        & Dropout(0.5) \\
        \midrule
        \multirow{1}{*}{BiLSTM} & 48x96+1+1, 48x8, 2 Layers\\
        \midrule
        \multirow{3}{*}{Functional 2} &Linear  48x16, 1\\
        \cline{2-2}
        & Dropout (0.5) \\
        \cline{2-2}
        & Sigmoid \\
        \bottomrule
    \end{tabular}
\end{table}

\begin{table}[ht]
    \centering
    \scriptsize
    \caption{Performance of GenéLive!\ with DDC's conv-stack, mean $\pm$ SD over 10 trials.}
    \label{tab:convstack}
    \begin{tabular}{c c | c c c}
        \toprule
        Dataset & Difficulty & F$_1$-score$^{\textrm m}$ & F$_1$-score$^{\textrm c}$ \\
        \midrule
        \multirow{5}{*}{Love Live!~All Stars}
        & Beginner & $0.8545 \pm 0.0040$ & $0.8535 \pm 0.0038$ \\
        & Intermediate & $0.7845 \pm 0.0040$ & $0.7852 \pm 0.0037$ \\
        & Advanced & $0.7758 \pm 0.0059$ & $0.7758 \pm 0.0064$ \\
        & Expert & $0.7777 \pm 0.0104$ & $0.7799 \pm 0.0107$ \\
        & all & $0.7909 \pm 0.0037$ & $0.8003 \pm 0.0034$ \\
        \midrule
        \multirow{6}{*}{Utapri}
        & Beginner & $0.6739 \pm 0.0113$ & $0.6714 \pm 0.0126$ \\
        & Intermediate & $0.7778 \pm 0.0052$ & $0.7719 \pm 0.0059$ \\
        & Advanced & $0.7584 \pm 0.0039$ & $0.7569 \pm 0.0042$ \\
        & Expert & $0.7917 \pm 0.0024$ & $0.7836 \pm 0.0028$ \\
        & Challenge & $0.8575 \pm 0.0040$ & $0.8575 \pm 0.0040$ \\
        & all & $0.7617 \pm 0.0048$ & $0.7383 \pm 0.0073$ \\
        \midrule
        \multirow{6}{*}{Stepmania}
        & Beginner & $0.7348 \pm 0.0337$ & $0.7247 \pm 0.0344$ \\
        & Intermediate & $0.7226 \pm 0.0070$ & $0.7190 \pm 0.0079$ \\
        & Advanced & $0.7394 \pm 0.0073$ & $0.7396 \pm 0.0064$ \\
        & Expert & $0.7321 \pm 0.0032$ & $0.7285 \pm 0.0051$ \\
        & Challenge & $0.7716 \pm 0.0036$ & $0.7683 \pm 0.0028$ \\
        & all & $0.7413 \pm 0.0045$ & $0.7282 \pm 0.0122$ \\
        \bottomrule
    \end{tabular}
\end{table}

\section{Evaluation on Fuzzy Label}\label{sec:fuzzy-label}
\Cref{tab:fuzzy} shows the evaluation result when the fuzzy label is not applied to label data during training.
The experimental results reproduced the results of the previous study \citep{Liang:2019} that \emph{fuzzy label} improves performance on imbalanced datasets.

\begin{table}[ht]
    \scriptsize
    \centering
    \caption{Performance of GenéLive!\ without the fuzzy label, mean $\pm$ SD over 10 trials.}
    \label{tab:fuzzy}
    \begin{tabular}{c c | c c}
        \toprule
        Dataset & Difficulty & F$_1$-score$^{\textrm m}$ & F$_1$-score$^{\textrm c}$ \\
        \midrule
        \multirow{5}{*}{Love Live!~All Stars}
        & Beginner & $0.8542 \pm 0.0044$ & $0.8527 \pm 0.0041$ \\
        & Intermediate & $0.7856 \pm 0.0045$ & $0.7857 \pm 0.0046$ \\
        & Advanced & $0.7780 \pm 0.0028$ & $0.7779 \pm 0.0028$ \\
        & Expert & $0.7763 \pm 0.0050$ & $0.7782 \pm 0.0049$ \\
        & all & $0.7922 \pm 0.0025$ & $0.8005 \pm 0.0029$ \\
        \bottomrule
    \end{tabular}
\end{table}

\section{Hyperparameters}\label{sec:hyperparametersAppendix}

\Cref{tab:hyperparameters} shows the list of hyperparameters of our model,
and the optimal value of each hyperparameters obtained by grid search.
The table also contains candidates of the grid search.

\begin{table}[ht]
    \centering
    \scriptsize
    \caption{List of hyperparameters.}
    \label{tab:hyperparameters}
    \begin{tabular}{c | c c}
        \toprule
        Name & Optimal value & Candidates \\
        \midrule
        learning rate start & $9\times10^{-4}$ & $1\times10^{-6}$, $2\times10^{-6}$, $4\times10^{-6}$, $9\times10^{-4}$ \\
        learning rate end & $9\times10^{-4}$ & $9\times10^{-4}$, $1\times10^{-3}$, $2\times10^{-3}$, $4\times10^{-3}$ \\
        $\eta_\textrm{min}$ & $1\times10^{-6}$ & $1\times10^{-6}$, $2\times10^{-4}$ \\
        fuzzy label scale & 0.2 & 0.1, 0.2, \dots, 0.7 \\
        fuzzy label width & 5 & 2, 3, 4, 5, 6 \\
        dropout (linear) & 0.3 & 0.3, 0.4, 0.5 \\
        dropout (RNN) & 0.1 & 0, 0.1, 0.2, 0.3 \\
        BCE loss weight & 64 & 48, 64, 96 \\
        \bottomrule
    \end{tabular}
\end{table}

\section{Selection of Max-pooling Kernel Size}\label{sec:maxpooling-kernel-size}
\begin{figure}[ht]
    \centering
    \includegraphics[width=\linewidth]{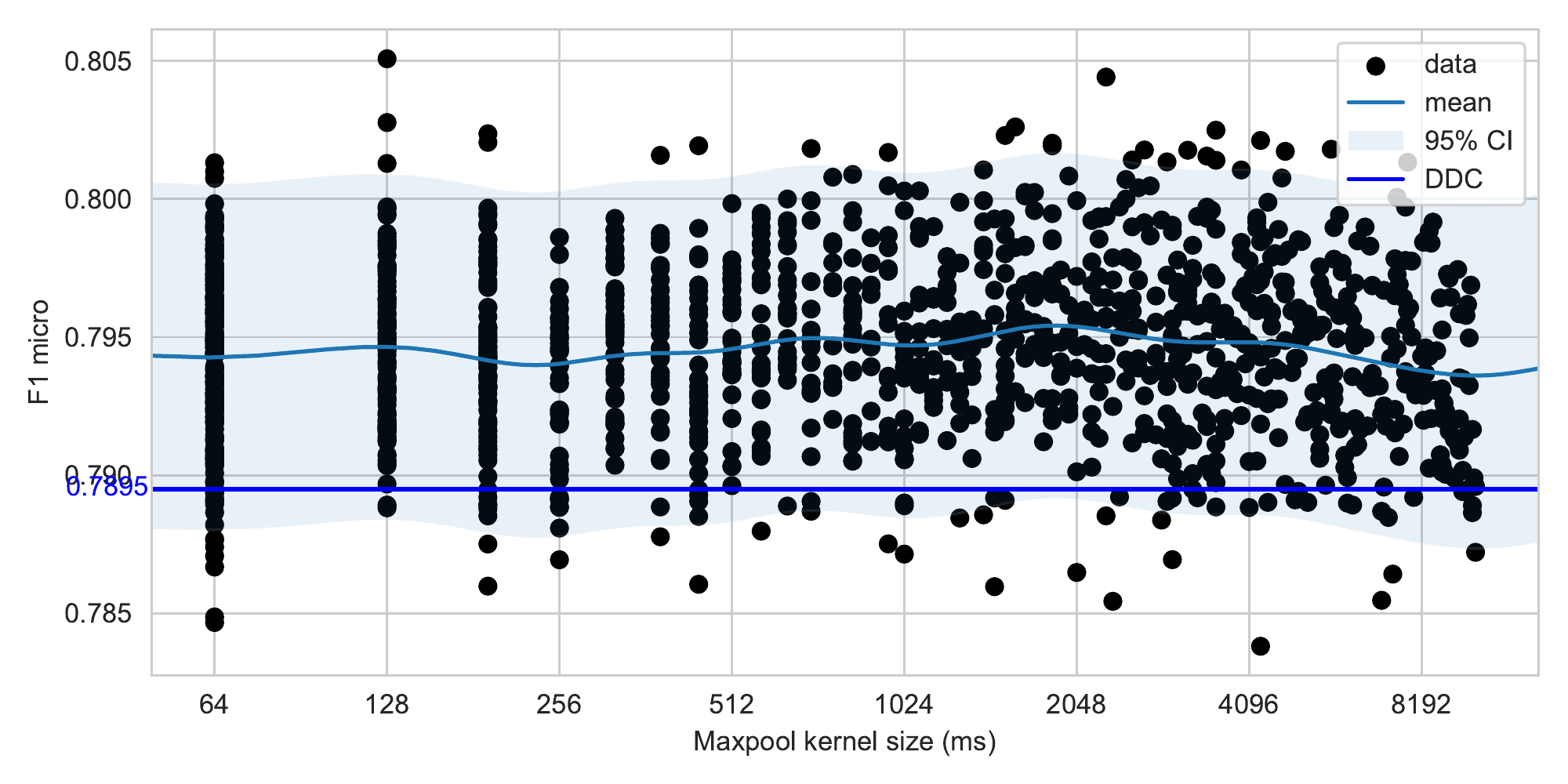}
    \caption{The performance when combining two conv-stacks one of whose max-pooling kernel size is 32 ms and varied the kernel size of the other.}
    \label{fig:max-pooling}
\end{figure}

As a preliminary experiment for multi-scale conv-stack described in \cref{sec:conv-stack}, we studied an effect of max-pooling kernel size which is appended at the side of the DDC conv-stack.
In the experiment, we sampled integer max-pooling kernel size from a log-uniform distribution between 64 and 10240 where the minimum unit is 64.
\Cref{fig:max-pooling} shows the result of the experiment.
The result shows that whatever the kernel size is, the performance of the model improves by a certain amount from DDC.
This suggests a hypothesis that if we append another conv-stack in parallel with different max-pooling kernel sizes, the performance of the model further improves.
To test the hypothesis, we tried multiple combinations of different max-pooling kernel sizes and ended up finding that only when a combination is meaningful to the task, the performance significantly improves (see \cref{sec:conv-stack}).

\section{Data Augmentation}\label{sec:dataAugmentationDetails}
We augment the audio of each song in the datasets.
The audio is first converted to a Mel spectrogram, which is a 2D array of time-frequency bins denoted by $S_0(t,f)\in\mathbb R^{TF}$, where $t=1,\dots,T$ denotes the $t$th time bin, and $f=1,\dots,F$ denotes the $f$th frequency bin.
The spectrogram $S_0(t,f)$ is then augmented via a series of transformations adopted from \citep{Park_2019}, resulting in an augmented Mel spectrogram $S$, which is an input to generative models.
We applied the following transformations in the presented order.
Note that each transformation is triggered by a predefined probability and applied to random positions $(t,f)$ of the spectrogram (see \texttt{loader\_aug\_config.yaml} for hyperparameters).
\begin{itemize}
    \item \emph{Frequency shift} shifts all frequency bins by a random amount $\delta_{t}$ drawn for each time bin $t$ from the uniform distribution on the interval of predetermined length $2\lambda$:
    \begin{equation}
        \begin{split}
            S_1(t, f) &= 
            \begin{cases}
                S_0(t, f - \delta_tF) & \text{$f > \delta_tF$}, \\
                S_0(t, F + f + 1 - \delta_tF) & \text{otherwise},
            \end{cases}\\
            &\text{ where } \delta_t \sim \mathcal U(-\lambda, \lambda).
        \end{split}
    \end{equation}

    \item \emph{Frequency mask} fills the $f$th frequency bins with the mean value for randomly chosen $f=f_1,\dots,f_k$:
    \begin{equation}
        S_2(t, f) = \frac{1}{TF}\sum_{i=1}^{T}\sum_{j=1}^{F} S_1(i, j).
    \end{equation}

    \item \emph{Time mask} fills the $t$th time bins with mean value for randomly chosen $t=t_1,\dots,t_k$:
    \begin{equation}
        S_3(t, f) = \frac{1}{TF}\sum_{i=1}^{T}\sum_{j=1}^{F} S_2(i, j).
    \end{equation}

    \item \emph{High frequency mask} also fills the $f$th frequency bins but such $f$ must be above the predetermined threshold $h$, preserving low frequency bins:
    \begin{equation}
        S_4(t, f) = 
        \begin{cases}
            \frac{1}{TF}\sum_{i=1}^{T}\sum_{j=1}^{F} S_3(i, j) & \text{$f \geq h$}, \\
            S_3(t, f) & \text{otherwise}.
        \end{cases}
    \end{equation}

    \item \emph{Frequency flip} reverses the order of frequency bins:
    \begin{equation}
        S_5(t, f) = S_4(t, F-f+1).
    \end{equation}

    \item \emph{White noise} adds a Gaussian noise, drawn for each $t, f$ with predetermined standard deviation $\sigma$:
    \begin{equation}
        \begin{split}
            S_6(t, f) &= S_5(t, f) + \delta_{t,f},\\
            &\text{ where } \delta_{t,f} \sim \mathcal N(0,\sigma^2).
        \end{split}
    \end{equation}

    \item \emph{Time stretch} stretches time bins to become $c$ times longer than the original length where $c$ is drawn from the uniform distribution on interval $[a,b]$ with predetermined $0 < a < b \le 1$:
    \begin{equation}\label{eq:augmentation}
        \begin{split}
            S(t, f) &= S_6(\lceil ct\rceil, f),\\
            &\text{ where } c \sim \mathcal U(a,b).
        \end{split}
    \end{equation}
\end{itemize}

\section{Full-Length Chart}\label{sec:chart_full}
\begin{figure}[ht]
    \centering
    \includegraphics[width=0.85\linewidth,trim=0 200 0 270]{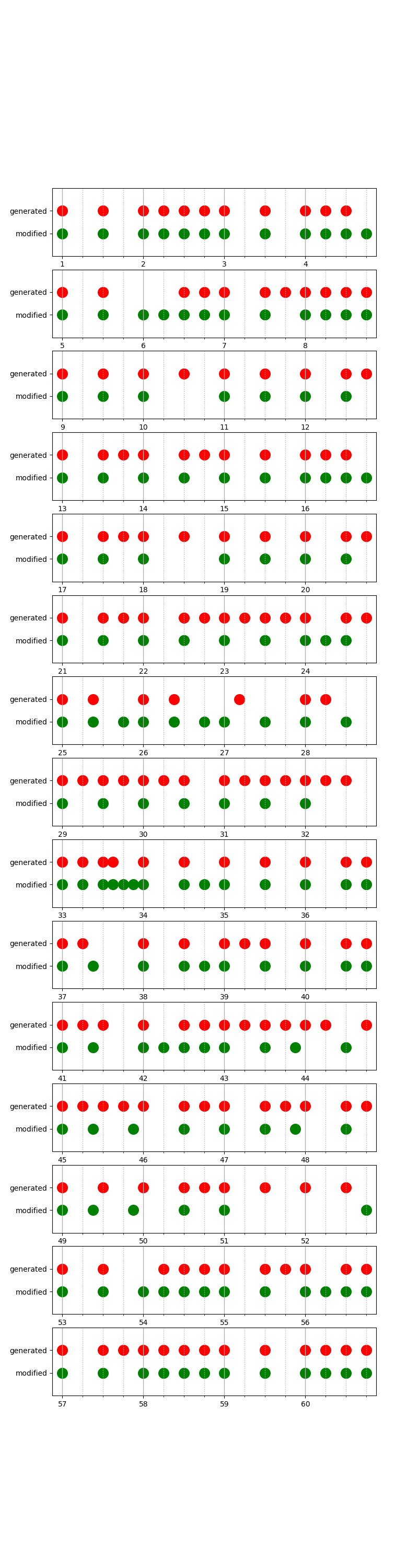}
    \caption{A generated chart for the Advanced difficulty mode and its manually-modified version (the whole 60 bars of the song).}
    \label{fig:chart-full}
\end{figure}

\Cref{fig:chart-full} shows the full-length comparison of the generated chart and the manually modified (released in production) chart, which we have seen in the first 8 bars in \cref{fig:chart}.
This song has 60 bars going on the following musical structure: its 8-bar length intro goes 2 repetitions of a 4-bar phrase; the 16-bar length verse goes 2 repetitions of an 8-bar phrase; the 10-bar length bridge goes a 4-bar phrase, 2 repetitions of a 1-bar phrase, and 2 repetitions of a 2-bar phrase; the 18-bar length chorus goes 2 repetitions of a 4-bar phrase, 2 repetitions of another 4-bar phrase, and a 2-bar extension of the last phrase; the 8-bar length outro repeats the same phrase as the intro.
We can see from the figure that GenéLive!~generates almost identical note patterns on such repetitions, which implies our multi-scale conv-stack almost correctly captures the repetition of 1-bar to 4-bar length phrases.

\Cref{tab:chart-full} shows the confusion matrix of those charts.
The generated chart has 166 notes whereas the manually-modified chart has 147 notes.
The precision, $\mathrm{TP}/(\mathrm{TP}+\mathrm{FP}) = 119/(119+47) \approx 0.72$, implies the artist team adopted 72\% of generated notes for the manually-modified (released) version, which means the generated chart gives a good draft that the artists can adopt the majority of generated rhythm patterns and can be inspired to improve the chart quality with their creativity.
The recall, $\mathrm{TN}/(\mathrm{TN}+\mathrm{FN}) = 294/(294+28) \approx 0.91$, implies only 9\% of the human-preferred rhythm patterns is missing; in other words, the generated chart contains a vast variety of desirable rhythm pattern candidates.
The accuracy, $(\mathrm{TP}+\mathrm{TN})/(\mathrm{TP}+\mathrm{FN}+\mathrm{FN}+\mathrm{TN}) = (119+294)/(119+28+47+294) \approx 0.85$, means the generated chart provides such good inspirations with little manual modification cost.

\begin{table}[ht]
    \footnotesize
    \centering
    \caption{Confusion matrix for the generated vs manual charts shown in \cref{fig:chart-full} (quantized by 8th notes).}
    \label{tab:chart-full}
    \begin{tabular}{cc|rr|r}
        & & \multicolumn{2}{c|}{Generated} &\\
                                &       & Positive & Negative & Probability \\
        \midrule
        \multirow{2}{*}{Manual} & Positive  &  119 &   28 & 0.81 \\
                                & Negative  &   47 &  294 & 0.86 \\
        \midrule
                                & Probability & 0.72 & 0.91 & 0.85 \\
    \end{tabular}
\end{table}

\section{Reproducibility Checklist}
\paragraph{
(1)
This paper includes a conceptual outline and/or pseudocode description of AI methods introduced.
}
Yes.
It is described in \cref{sec:proposed-method}.

\paragraph{
(2)
This paper clearly delineates statements that are opinions, hypothesis, and speculation from objective facts and results.
}
Yes.
It is described in \cref{sec:introduction}.

\paragraph{
(3)
This paper provides well marked pedagogical references for less-familiar readers to gain background necessary to replicate the paper.
}
Yes.
It is described in \cref{sec:introduction,sec:related-work}.

\paragraph{
(4)
Does this paper make theoretical contributions? If your answer is Yes, answer questions 15-21. If your answer is No, go to question 22 directly.
}
No.

\paragraph{
(5)
Does this paper rely on one or more data sets? If your answer is yes, answer questions 23-28. If your answer is no, go to question 29 directly.
}
Yes.

\paragraph{
(5.1)
A motivation is given for why the experiments are conducted on the selected datasets.
}
Yes.
It is described in \cref{sec:introduction}.

\paragraph{
(5.2)
All novel datasets introduced in this paper are included in a data appendix.
}
No.
Our novel datasets, ``Love Live!~All Stars'' and ``Utapri,'' are not included in the data appendix for copyright reasons.
We are not a copyright holder of those songs and licensed for internal use only.

\paragraph{
(5.3)
All novel datasets introduced in this paper will be made publicly available upon publication of the paper with a license that allows free usage for research purposes.
}
No.
Our novel datasets, ``Love Live!~All Stars'' and ``Utapri,'' will not be made publicly available for copyright reasons.
We are not a copyright holder of those songs and licensed for internal use only.

\paragraph{
(5.4)
All datasets drawn from the existing literature (potentially including authors’ own previously published work) are accompanied by appropriate citations.
}
Yes.
The open dataset ``Stepmania'' is cited in \cref{sec:data-acquisition}.

\paragraph{
(5.5)
All datasets drawn from the existing literature (potentially including authors’ own previously published work) are publicly available.
}
Yes.
The open dataset ``Stepmania'' is available at DDC's code repository \url{https://github.com/chrisdonahue/ddc}.
Note that our ``Stepmania F'' and ``Stepmania ITG'' refer to DDC's ``Flaxtil'' and ``ITG\@,'' respectively, and our ``Stepmania'' is the union of them.

\paragraph{
(5.6)
All datasets that are not publicly available are described in detail, with explanation why publicly available alternatives are not scientifically satisficing.
}
Yes.
It is described in \cref{sec:introduction}.

\paragraph{
(6)
Does this paper include computational experiments? If your answer is Yes, answer questions 30-41. If your answer is No, you can skip the rest of the questions.
}
Yes.

\paragraph{
(6.1)
Any code required for pre-processing data is included in the appendix.
}
Yes.
All the preprocessing code is included in \texttt{notes\_generator/preprocessing/} directory in the supplementary material.

\paragraph{
(6.2)
All source code required for conducting and analyzing the experiments is included in a code appendix.
}
Yes.
We made figures in the paper by using \texttt{figure.ipynb} file in the supplementary material.

\paragraph{
(6.3)
All source code required for conducting and analyzing the experiments will be made publicly available upon publication of the paper with a license that allows free usage for research purposes.
}
Yes.
We will publish all the source code (except for the part depending on proprietary datasets) on GitHub under the MIT license.

\paragraph{
(6.4)
All source code implementing new methods have comments detailing the implementation, with references to the paper where each step comes from.
}
Yes.
Since our implementation is deployed in real game products, it contains a lot of production-related code (even after we have removed most of them).
Thus, it is impractical to add comments on all lines of code.
Instead, we put remarks on which files implement the core of our method, i.e., the beat guide and multi-scale conv-stack, and selectively added comments in those files with appropriate references to the paper.
The remarks are found in \texttt{README.md} file in the supplementary material.

\paragraph{
(6.5)
If an algorithm depends on randomness, then the method used for setting seeds is described in a way sufficient to allow replication of results.
}
No.
In our experiments, the current time was used as a seed for the following reasons.
As noted in the PyTorch document\footnote{\url{https://pytorch.org/docs/stable/notes/randomness.html}}, our multi-GPU experiments cannot be replicated even when using identical seeds.
Instead of fixing seeds, we ensure reproducibility in the statistical sense.
All the results in the paper and appendix have very small standard deviations (10--100x smaller than averages).
Thus, one can reproduce almost identical results whatever seeds and hardware are used.

\paragraph{
(6.6)
This paper specifies the computing infrastructure used for running experiments (hardware and software), including GPU/CPU models; amount of memory; operating system; names and versions of relevant software libraries and frameworks.
}
Yes.
See \cref{sec:training-methodology}.

\paragraph{
(6.7)
This paper formally describes evaluation metrics used and explains the motivation for choosing these metrics.
}
Yes.
See \cref{sec:training-methodology}.

\paragraph{
(6.8)
This paper states the number of algorithm runs used to compute each reported result.
}
Yes.
See \cref{sec:training-methodology}.

\paragraph{
(6.9)
Analysis of experiments goes beyond single-dimensional summaries of performance (e.g., average; median) to include measures of variation, confidence, or other distributional information.
}
Yes.
All averages or medians in the result tables are accompanied by standard deviation, and all result figures show the scatter plot of individual trials or averages/medians with confidence intervals.

\paragraph{
(6.10)
The significance of any improvement or decrease in performance is judged using appropriate statistical tests (e.g., Wilcoxon signed-rank).
}
Yes.
Our main results shown in \cref{tab:result} are tested by the Wilcoxon signed-rank test.

\paragraph{
(6.11)
This paper lists all final (hyper-)parameters used for each model/algorithm in the paper’s experiments.
}
Yes.
See \cref{tab:hyperparameters}.

\paragraph{
(6.12)
This paper states the number and range of values tried per (hyper-) parameter during development of the paper, along with the criterion used for selecting the final parameter setting.
}
Yes.
See \cref{tab:hyperparameters}.

\end{document}